%% file: survey.tex
\documentclass[manuscript,nonacm]{acmart}
\AtBeginDocument{%
  }

\usepackage{lscape}
\usepackage{subfigure}
\usepackage[normalem]{ulem}
\usepackage{todonotes}
\usepackage{comment}
\usepackage{float}

\begin{document}

\title{Generative Artificial Intelligence for Internet of Things Computing: \\A Systematic Survey}

\author{Fabrizio Mangione}
\authornote{All authors contributed equally to this work. We acknowledge the support of the  NextGenerationEU Project "Entrust: usEr ceNtric plaTform foR continoUS healThcare"–CUP H53D23008110001- 3, under the Italian MUR program. We also acknowledge the support of the PNRR project FAIR - Future AI Research (PE00000013), Spoke 9 - Green-aware AI, under the NRRP MUR program funded by the NextGenerationEU. }
\email{fabrizio.mangione@dimes.unical.it}
\orcid{https://orcid.org/0009-0000-2581-0796}
\affiliation{%
  \institution{ University Of Calabria}
  \city{Rende}
  \country{Italy}
}

\author{Claudio Savaglio}
\affiliation{%
  \institution{University Of Calabria}
  \city{Rende}
  \country{Italy}}
\email{csavaglio@dimes.unical.it}
\orcid{https://orcid.org/0000-0001-5092-0823}

\author{Giancarlo Fortino}
\affiliation{%
  \institution{ University Of Calabria}
  \city{Rende}
  \country{Italy}}
\email{giancarlo.fortino@unical.it}
\orcid{https://orcid.org/0000-0002-4039-891X}

\renewcommand{\shortauthors}{Mangione et al.}
\renewcommand{\shorttitle}{Generative AI for IoT Computing: A Systematic Survey}
\begin{abstract}
The integration of Generative Artificial Intelligence (GenAI) and the Internet of Things (IoT) is attracting significant attention due to their rapid, individual evolution and their widespread joint adoption across domains such as Healthcare, Manufacturing, and Smart Cities. Consequently, their increasing popularity has catalyzed extensive research to understand the interplay between GenAI and IoT, and how their synergy can advance the state-of-the-art in both domains. However, despite the increasing prominence of GenAI for IoT Computing, much of the existing research remains focused on specific, narrowly scoped applications. This fragmented approach highlights the need for a more comprehensive analysis of the potential, challenges, and implications of GenAI integration within the broader IoT ecosystem. The survey exactly aims to address this gap by providing a holistic overview of the opportunities and issues arising from the convergence of these mainstream paradigms. Our contribution is 
a systematic literature review following the PRISMA methodology: a comparison framework is presented, and well-defined research questions are outlined to comprehensively explore the current and future directions of GenAI-aided  IoT Computing.
\end{abstract}

\ccsdesc[500]{Computing methodologies~Artificial intelligence}
\ccsdesc[500]{Computing methodologies~Machine learning}
\ccsdesc[500]{Computer systems organization~Embedded and cyber-physical systems}

\maketitle
\input{survey_sections/introduction}

\input{survey_sections/background}

\input{survey_sections/research_methodology}

\input{survey_sections/literature_review}
\input{survey_sections/practical_challenges_of_generative_ai_for_iot_systems}

\input{survey_sections/conclusion}
\bibliographystyle{plain}
\bibliography{survey}
\end{document}

%% file: survey_sections/introduction.tex
\section{Introduction}
\label{sec:introduction}

According to more recent industry forecasts, the global number of connected \textbf{Internet of Things (IoT)} devices is expected to continue to rise well into the next decade, exceeding 40 billion by 2030 \cite{statista2024iot,9210375}. 
Looking further ahead, some analysts anticipate that by 2035 IoT deployments, especially at the network's edge, could generate thousand of zettabytes of data annually, representing a significant share of the overall global datasphere \cite{onio2024edge, huawei2021intelligent}. 
These projections underscore the growing importance of effective computing strategies. As the IoT ecosystem scales, massive volumes of devices and data must be managed to provide advanced cyber-physical services. Simultaneously, this management must ensure security, reliability, usability, and efficiency across increasingly complex networks \cite{8972389,NIZETIC2020122877}.
However, current IoT solutions  face significant hurdles in dealing simultaneously with scalability, data heterogeneity, and resource constraints \cite{cheikh2025energyscalabilitydatasecurity} \cite{9475501}.

In this context, \textbf{Generative Artificial Intelligence (GenAI)} 
has emerged as a paradigm with capabilities extending significantly beyond traditional AI and earlier iterations of Edge Artificial Intelligence (Edge AI) \cite{bdcc7010044}. 
While traditional AI focuses on analytical and predictive tasks \cite{bishop2007,cao2023comprehensivesurveyaigeneratedcontent} and Edge AI emphasizes on-device intelligence \cite{bdcc7010044}, GenAI  leverages the principles of \textbf{Deep Generative Models (DGMs)} to synthesize new contents. Catalyzing ever-increasing interest, in a few years, it has evolved from a framework for uncovering data distributions into a versatile suite capable of generating data, code, and complex reasoning \cite{10234662,wei2023chainofthoughtpromptingelicitsreasoning}, 
 leading to the development of novel and multidisciplinary research fields.
In particular, GenAI represents a paradigm shift fundamentally altering how data is managed and utilized within the IoT ecosystem. By generating synthetic data and optimizing interpretation, GenAI tackles inherent IoT challenges such as data accuracy, uncertainty, human interaction, and system reliability in resource-constrained environments. Specifically, it facilitates the creation of realistic scenarios and enhances predictive models, while supporting autonomous decision-making and real-time human-machine interaction.
Despite this potential, the integration of GenAI into \textbf{IoT Computing} (defined as the paradigm encompassing data-, thing-, and network- related aspects of the ecosystem), remains a complex endeavor. 
Indeed, the explosive and widespread adoption of the GenAI term has led to two, 
misconceptions that hinder rigorous research: \textit{(i)} an oversimplification of GenAI, wherein AI solutions are incorrectly categorized under this label, and \textit{(ii)} 
the conflation of specific architectures (e.g., Large Language Models and Generative Adversarial Networks) with the broader field of GenAI. Both these misconceptions have already contributed to the mischaracterization of GenAI in many different research areas \cite{barassi2024toward}. On a technical level, this may obscure the fundamental differences in model architectures, training objectives, and operational requirements;  on a regulatory level, this risks imposing undue restrictions or ethical concerns on AI that do not share the same risks and implications as DGMs.
Applying these misconceptions to the IoT domain can lead to even deeper conceptual distortions and hinder a proper understanding of the novel opportunities and challenges emerging from their convergence.
While the body of literature regarding AI in IoT is vast, there are few surveys specifically targeting the intersection of GenAI and IoT Computing. As highlighted in Tab. \ref{tab:Comparison framework for the analyzed surveys}, existing studies  mainly date back to the last two years and they often focus on vertical application domains (especially networking) or lack a systematic methodology, limiting their reproducibility and scope. Furthermore, there is a lack of comprehensive analysis that rigorously examines the theoretical underpinnings, architectural requirements, and cross-domain applications of this convergence.
To bridge this gap, this article provides a \textit{comprehensive},  \textit{systematic}, and \textit{cross domain} survey of GenAI for IoT Computing, distinguishing itself through a broader scope and a more rigorous investigation. Spanning literature from 2020 to 2026 and conducted in accordance with the PRISMA methodology \cite{PAGE2021103}, we analyze \textcolor{black}{78} eligible studies to provide a robust comparison framework for this emerging field. 
The main contributions of this article can be summarized as follows:
\begin{itemize}
    \item \textbf{Systematic Taxonomy:} We provide a structured taxonomy of DGM, specifically tailored for IoT environments and enabled GenAI applications, aimed at clarifying key features and common architectural misconceptions.
        \item \textbf{Comprehensive State-of-the-Art Analysis:} We exhaustively explore the current literature, classifying contributions based on their application to device, network and data management.
        \item \textbf{Practical Constraints and Mitigation Strategies:} We analyze the practical constraints hindering the integration of GenAI in IoT, accompanied by a discussion of strategies to mitigate these issues.
\end{itemize}

\begin{table}[h!]
\caption{Overview on the similar existing surveys.}
\label{tab:Comparison framework for the analyzed surveys}
\begin{center}
\footnotesize
\begin{tabular}{p{.8cm} p{.8cm} p{2.5cm} p{5.8cm} p{1.5cm} p{1.5cm}}
\hline

\toprule

\textbf{Ref.} & \textbf{Year} & \textbf{App. Domain} & \textbf{Scope} & \textbf{Systematic} & \textbf{\# articles} \\ \midrule

\cite{vu2024applications} & 2024 & Networking & Application of GDM for mobile wireless networking & No & 27 \\ \midrule

\cite{Navidan_2021} & 2021 & Networking & Application of GenAI (GANs) in networking & No & 54 \\ \midrule

\cite{10422716} & 2024 & Networking & GenAI support for IoT applications in 6G networks & No & 10\\ \midrule

\cite{chen2024generativeaidrivenhumandigital} & 2024 & Healthcare & GenAI-driven Human Digital Twin in IoT healthcare & Yes & 172\\ \midrule
 
\cite{10623653} & 2024 & Autonomous systems & Analyze the role of GenAI in Autonomous Systems & No & 17\\ \midrule

\cite{10669603} & 2024 & Edge AI & Deployment and integration of LLMs at the IoT edge  & No & 16\\ \midrule

\cite{xu2024unleashing} & 2023 & Edge AI & Analyze the deployment of DGMs at the IoT edge & No & 72\\ \midrule

\cite{ALWAHEDI2024167} & 2024 & IoT security & GenAI for intrusion and anomaly detection in IoT & Yes & 50\\ \midrule

\textcolor{black}{\cite{11125750} }& 2025 & \textcolor{black}{Smart City} &\textcolor{black}{ GenAI to make complex data accessible to humans} & No & 60\\ \midrule

\textbf{Our} 
& \textbf{2026} & \textbf{Cross domain} & \textbf{GenAI \& IoT Computing convergence} 
& \textbf{Yes} & \textbf{78} \\ \bottomrule

\end{tabular}
\end{center}
\end{table}

 Thus, this article seeks to provide a comprehensive and insightful survey, delving into the theoretical underpinnings, architectural frameworks, enabling technologies, applications and challenges arising from the exploitation of GenAI for IoT Computing. In particular, the remainder of the manuscript is organized as follows. \textbf{Section~\ref{sec:background}} introduces basic concepts behind GenAI and IoT Computing, offering a solid conceptual foundation
. In \textbf{Section~\ref{sec:research methodology}} is provided a detailed report about the pursued research objectives and adopted PRISMA-based search methodology. Then, in \textbf{Section~\ref{sec:literature_review}} is presented a systematic review of the existing literature, conducted using a theoretical framework which synthesizes the current research landscape and provides answers to the outlined research questions. 
\textbf{Section~\ref{sec:challenges}} delves into cross-cutting solutions for main  practical challenges hindering the exploitation of GenAI for IoT Computing. 
Lastly, \textbf{Section~\ref{sec:conclusion}} concludes the article by summarizing the key findings and open issues.

%% file: survey_sections/background.tex
\section{Background}
\label{sec:background}

\subsection{Towards Intelligent Internet of Things Computing}
\label{sec:IoT}
The IoT defies a straightforward definition due to its expansive and evolving nature, which encompasses diverse technologies, applications, and use cases \cite{9306923}. At its core, IoT refers to a network of uniquely identifiable physical or virtual entities, known as \textit{Things}, equipped with sensors, actuators, and embedded computing. These \textit{Things} communicate, exchange data, and act autonomously or semi-autonomously, bridging the physical and digital realms to create adaptive environments hosting  context-aware processes \cite{8241454}. IoT systems range in complexity from basic implementations to large-scale, self-managed ecosystems, 
enabling ubiquitous connectivity, interoperable communication, self-management and sophisticated service delivery  \cite{CASADEI2025101548}. 
Given such complexity and versatility, it is not surprising that
the interpretation of IoT varies significantly among stakeholders, such as businesses, academic researchers, and standardization bodies, each adopting perspectives shaped by their priorities and expertise. 

Regardless of the specific interpretation or domain of interest, however, the IoT Computing (intended as the paradigm dealing with all the broader computational aspects within IoT systems, including not only the software but also the data, architectural and infrastructural aspects) 
 encompasses three main dimensions  \cite{ATZORI20102787}: \textit{Internet}, \textit{Things}, and \textit{Semantic} one.
The \textbf{Internet-Centric perspective} emphasizes robust connectivity and advanced communication technologies, which are essential for linking diverse devices into a global infrastructure. The focus is on standardized protocols and innovative networking solutions ensuring interoperability, scalability, and reliability across heterogeneous systems. Differently, the \textbf{Things-Centric perspective} highlights the \textit{Things} and the technologies required for the provision of  the cyber-physical services enabling novel forms of interplay between the real and the virtual worlds as well as among humans and (smart) objects. Finally, 
the \textbf{Semantic-Centric perspective} addresses the challenge of ensuring data relevance and usability, leveraging semantic frameworks to establish intelligent systems capable of understanding relationships and context within vast datasets. This facilitates adaptability, situational awareness, and informed decision-making. Integrating these three dimensions transforms IoT Computing into a holistic framework that seamlessly bridges physical and digital systems, also enhancing the overall scalability, robustness, and intelligence. 

Driven by the shift toward Intelligent IoT Computing, since 2012 AI has been leveraged to 
analyze in real-time vast volumes of sensor data, facilitate the optimization of processes, devices and networks, enhance the user experiences, and improve operational efficiency across multiple sectors, including Healthcare, Smart Cities, and Industry 4.0 \cite{wu2025aigenerationsai10}. Where to deploy AI to engineer intelligent IoT systems opened, however, another  discussion. Initially, \textbf{Cloud} services have been instrumental in supporting this convergence by providing the computational capacity 
required for training sophisticated AI models and storing extensive IoT data streams. However, exclusive reliance on Cloud services introduces significant network latency, efficiency and privacy issues, which presents a critical bottleneck for several IoT applications \cite{202502.1500}. To mitigate these challenges, the paradigm of \textbf{Edge AI}~\cite{bdcc7010044,electronics14244877} has emerged as a compelling solution. Edge AI entails deploying AI capabilities closer to the data source, at the network's periphery, thereby facilitating more rapid data processing and decision-making. By performing computations locally, Edge AI reduces the need to transmit data to distant cloud servers, significantly diminishing latency while simultaneously enhancing data privacy and security. Notwithstanding its advantages, Edge AI is not without its limitations. The computational and energy resources available at edge nodes are typically far more constrained than those in centralized cloud environments, thereby limiting the complexity and scale of the AI models that can be executed at the edge \cite{ali2025edgeartificialintelligencesystematic}. Additionally, maintaining distributed intelligence across heterogeneous edge devices poses substantial challenges in terms of coordination, scalability, and the continuous updating of AI models. 
As such, for harnessing the full potential of AI-enabled IoT ecosystems is of paramount importance: \textit{(i)} to achieve an effective balance between cloud and edge processing to spread intelligence in the computing continuum, and \textit{(ii)} to propose novel solutions to mitigate inherent IoT challenges related to poor data quality, not intuitive human interaction, complex device management, and unpredictable system reliability \cite{electronics9122083,s20226441}. 
Exactly there, 
as explored in the rest of the article, \textbf{GenAI} possesses capabilities that distinguish it from both traditional AI and Edge AI (mostly exploiting deterministic approaches rather than probabilistic ones), thereby fostering a uniquely different contribution to IoT Computing for the engineering of intelligent systems \cite{wang2024internetthingseragenerative}.

\subsection{Deep Generative Models - DGMs}
Today, the terms DGMs and GenAI are frequently used interchangeably as synonyms. However, DGMs should be used when referring to a specific class of models, while GenAI should be used in a broader context to describe the practical implementation of DGMs and other techniques or multimodal architectures to create real-world applications.
Driven by rapid advancements in Neural Networks (NNs) and significant increases in computational power, DGMs has emerged as a central area of study within Machine Learning (ML) and AI, particularly in its ability to approximate complex data distributions and generate synthetic data that closely resemble real-world instances. Its applications span a broad spectrum of domains, 
encompassing classical ML modalities, such as Text analysis, Image analysis, and Audio analysis \cite{team2024gemini, hurst2024gpt}, 
to emerging IoT applications in areas such as Healthcare \cite{kung2023performance}, Intelligent Transportation Systems (ITS) \cite{10811280}, and Smart Cities \cite{li2024urbangpt}. 
As opposed to \textbf{Discriminative AI models}, which focus solely on partitioning and categorizing data points, DGMs attempt to capture the \textbf{underlying 
probabilistic distribution}, thereby facilitating the synthesis of new, high-fidelity examples. In fact, from a mathematical standpoint, the main objective of DGMs is to learn a representation of an unknown and often intractable probability distribution in a high-dimensional space. Rather than explicitly modeling the probability itself, the focus is on learning a generator that transforms samples from a simple and well-understood distribution (typically Gaussian noise) into data points that closely resemble the observed data. However, learning such a generator is generally impractical or infeasible for most real-world datasets, and even when it is possible, it remains a challenging task. Consequently, it has become standard practice to approximate the generator by using generic functions like NNs with multiple hidden layers. This approach forms the core design principle in DGMs, where the generator is represented by a feed-forward Deep Neural Network (DNN) \cite{jakub}. 
To fully grasp the potential of DGMs, two fundamental concepts must be considered: \textbf{uncertainty} and \textbf{understanding}. Consider a traditional AI classification task that categorizes objects into two classes. If a model is trained only to separate classes (the discriminative approach), it learns where to draw the boundary between them. When a new sample falls clearly on one side of that boundary, the model confidently assigns it to a class, even if the sample looks very unusual compared to what it has seen during training . When the model also learns how the data are distributed (the joint or generative perspective), it not only decides which class a sample belongs to, but also whether the sample looks plausible at all. In this case, a sample may be far from the boundary but still lie in a region where neither class typically appears. Practically, this allows the model to recognize that the input is unfamiliar or unlikely, leading to more cautious decisions, better uncertainty awareness, and improved handling of out-of-distribution or anomalous data \cite{bond2021deep, bishop2006pattern}.
So, albeit in the literature DGMs are frequently regarded primarily as mechanisms for generating new data that mimics human creativity, behavior, or appearance, we advocate for a broader perspective, where the real value lies in learning how data are distributed \cite{kingma2022autoencodingvariationalbayes}. 
This broader use of DGMs turns them into a core building block for realizing AI systems that are more reliable, robust, and trustworthy in real-world scenarios. 

The process through which DGMs capture the underlying probabilistic distribution of data is central to their ability to generate realistic outputs. This objective is accomplished using a variety of methodologies, each offering distinct approaches to learning and representing data distributions. Among the most prominent strategies are Adversarial Training (as utilized in GANs \cite{goodfellow2014generativeadversarialnetworks}) and Maximum Likelihood Estimation (MLE) \cite{sohl2015deep}, which forms the theoretical foundation for Diffusion-based methods. Both are widely adopted and impactful strategies in generative modeling. For the purpose of this discussion, we focus on DGMs that operate based on the principle of MLE (i.e., to choose the model parameters to maximize the probability assigned to the observed data, ensuring an accurate representation of the underlying distribution): indeed, MLE isn’t universal for DGMs, but many can adapt to it \cite{theis2015note}.
\subsection{Taxonomy of DGMs and IoT-related implications}
\label{sec:dgm}
\begin{figure}[ht!]
    \centering \includegraphics[width=.9\linewidth]
    {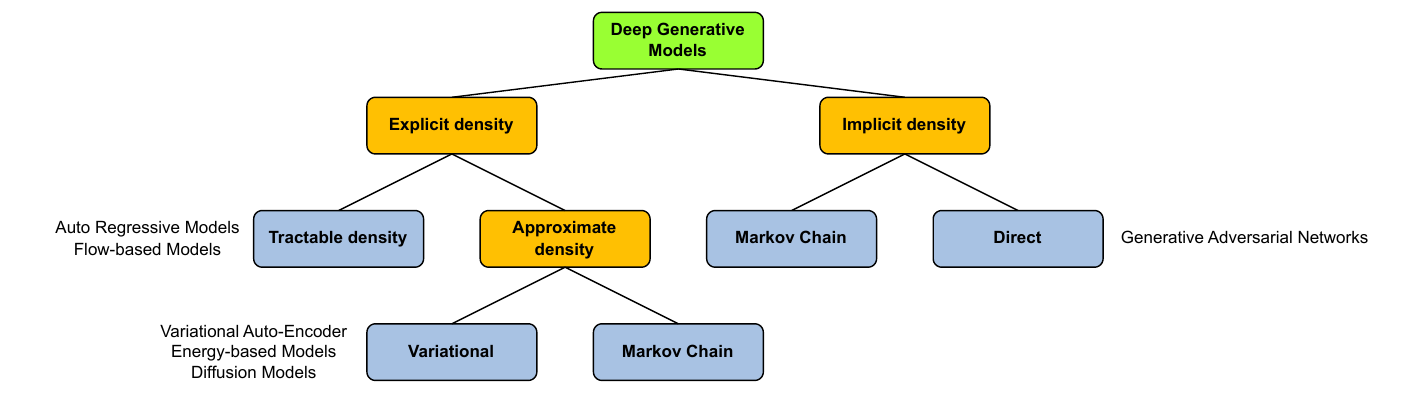}
    \caption{A taxonomy of DGMs inspired from \cite{goodfellow2017nips2016tutorialgenerative}: classes (boxes) and related instances}
    \label{fig:fig2}
\end{figure}

Having emphasized the critical importance and wide-ranging applicability of DGMs, in the following we examine how such models are formulated, according the taxonomy shown in Fig. \ref{fig:fig2} and presented in \cite{goodfellow2017nips2016tutorialgenerative}. The green rectangle represents the overarching category of DGMs, while the orange rectangles denote high-level subclasses that further branch into more specific model types (in light blue rectangles). Indeed, even based on the same MLE principles, DGMs differ in their approach to representing or approximating the likelihood. 
Specifically, Explicit density models (left branch of Fig.~\ref{fig:fig2}), construct a probability density providing an explicit likelihood that can be maximized. Within these models, the density may be computationally tractable, or it may require Variational or Monte Carlo approximations to optimize the likelihood. On the other hand, Implicit density models (right branch of Fig.~\ref{fig:fig2}) do not directly define a probability distribution over the data space. Instead, they interact indirectly with the distribution, often sampling from it. Some implicit models use Markov Chains to stochastically transform one sample into another, while others generate samples in a single step without input. Albeit GANs can theoretically define an explicit density, their training relies on solely on sampling, aligning them with implicit models that directly sample from the distribution.
Therefore, in the following, we provide a concise overview of the two main categories of DGMs, i.e.,  Explicit and Implicit density models, along with their respective subcategories. We particularly focused on their overall pros-and-cons with respect to IoT Computing, with main takeaways of this analysis summarized in Tab.~\ref{tab:KOC}. 

\subsubsection{Explicit density}
The primary challenge in explicit density models lies in designing architectures capable of capturing the full complexity of the data while preserving computational tractability \cite{jakub}. Two main strategies address this issue: \textit{(i)} constructing models with carefully designed structures that inherently ensure computational tractability, as exemplified by the class of Tractable density Models, and \textit{(ii)} employing models that allow for tractable approximations of the likelihood and its gradients, as represented by the class of Approximate density Models.
 The \textbf{Tractable density} class includes DGMs that explicitly define a probability density function, enabling direct likelihood evaluation. This category comprises Auto Regressive models, which decompose the joint distribution into sequential dependencies, and Flow-based models, which use invertible transformations to ease likelihood computation.

\paragraph{\textbf{Auto Regressive Models (ARMs)}}
These models represent the distribution of data sequentially, predicting each element based on the preceding ones. Modeling all possible conditional distributions individually is impractical, as it would require building a separate model for each element, with increasing complexity due to the growing number of conditions.
A potential solution to the issue is utilizing a single, shared model for the conditional distribution. The initial approach to reducing complexity involves assuming a finite memory constraint. Specifically, that each variable depends on no more than two other variables. 
This method holds particular relevance for IoT applications, which are frequently characterized by resource-constrained devices incapable of managing computationally intensive tasks. However, this approach has an obvious drawback that is the limited memory range. A possible solution to short-range memory is the application of Recurrent Neural Networks (RNNs), like the Long Short Term Memory (LSTM) \cite{LSTM}, that allow long-range dependency learning. This approach gives a single parametrization, thus it is efficient and also solves the problem of a finite memory. However, is sequential, hence slow, and due to the application of RNNs, problems like exploding or vanishing gradients need to be addressed. An alternative approach utilizes Convolutional Neural Networks (CNNs) rather than RNNs to model the probability distribution. This approach offers distinct advantages, specifically the use of shared kernels and improved process parallelization. However, this solution is inefficient when there is a need to sample a new object. CNNs are generally more computationally expensive and heavier than simple RNNs, especially in terms of parameter count, memory consumption and computation complexity, limiting their use in IoT applications. 
Another important class of models constitutes Transformers \cite{vaswani2023attentionneed}, which uses self-attention layers instead of causal convolutions. Like LSTMs, Transformer architectures are capable of handling distant information. 
Unlike LSTMs, Transformers do not rely on recurrent connections, making them easier to parallelize and more efficient. They consist of stacked Transformer blocks that map input sequences to output sequences of the same length. Each block combines linear layers, feed-forward networks, and, most importantly, self-attention. Self-attention captures contextual relationships by integrating information across tokens between layers, enabling progressively richer and more contextualized representations throughout the network.
In conclusion, the main advantage of ARMs is that they can learn long range statistics and, in a consequence, powerful density estimators. However, their drawback is that they are parameterized in an auto regressive manner, hence sampling is a slow process. Moreover, they lack a latent representation, therefore, it is not obvious how to manipulate their internal data representation. In IoT domains, the application of Transformer-based models have gained a considerable interest motivated by the enhanced data processing, thanks to the self attention mechanism, and the improved decision making, that leads to more adaptive and intelligent systems \cite{10.1145/3638550.3641126, 10731639}. However, IoT devices often have limited computational resources, making it challenging to deploy resource-intensive Transformer models. Furthermore, their effectiveness can be hindered by the limited availability of labeled data in specific IoT applications, affecting their performance \cite{10.1145/3638550.3641126}.
\paragraph{ \textbf{Flow-based Models}}
The change of variables formula provides a principled manner of expressing a density of a 
random variable by transforming it with an invertible transformation. 
The key point of Flow-based models, is to seek for such NNs
that are both invertible and the logarithm of the determinant of a Jacobian matrix is relatively easy to calculate. The resulting models, that consists of invertible transformation with tractable
determinant of a Jacobian matrix, are referred to as Normalizing Flows or Flow-based Models. 
Despite the success of Normalizing Flows models, in estimating high-dimensional densities, certain limitations persist in their design. 
First, the latent space onto which input data is projected is not lower-dimensional, meaning Flow-based models do not inherently support data compression and are computationally demanding, limiting their use in IoT applications. Flow-based models also exhibit notable challenges in estimating the likelihood of out-of-distribution samples, namely samples that originate from distributions different from the training set. This could represent a significant problem in IoT applications wherein there is a huge data variability \cite{ATZORI20102787}. One of the most compelling features of normalizing flows is the invertibility of their learned bijective mapping. This property arises from specific design constraints in models, which theoretically ensure invertibility. The integrity of the inverse map is crucial for the applicability of the change-of-variable theorem, accurate computation of the Jacobian, and reliable sampling. However, in practice, numerical errors can break invertibility, causing the inverse mapping to diverge \cite{10731639}.

\subsubsection{Approximate density} To bypass the 
design constraints of models with tractable density functions, alternative approaches have been developed: they retain explicit density functions, but accept intractability, necessitating the use of approximations for likelihood maximization. These models can be broadly categorized into two groups: those employing deterministic approximations, typically via variational methods, and those relying on stochastic approximations, commonly utilizing Markov Chain Monte Carlo (MCMC) techniques.

\paragraph{ \textbf{Variational Auto-Encoder}} 
The models in this group are based on the assumption of a lower-dimensional latent space coupled with an underlying generative process. In this context, latent variables represent hidden factors that explain the observed data, while the conditional relationship between latent variables and observations can be interpreted as a generator. For this reason, these approaches are commonly referred to as Latent Variable Models. A well-known example is probabilistic Principal Component Analysis (pPCA) \cite{0f9485aa-56d5-346a-95e4-aa1c6bc30fc3}, in which both the latent variables and the observed data are modeled using Gaussian distributions, and their relationship is assumed to be linear.
In IoT applications, data collected from heterogeneous sensors frequently involve intricate nonlinear correlations. Since pPCA assumes a linear dependency structure, it may fail to capture the underlying variability in such datasets making alternative nonlinear techniques more suitable.
A non-linear extension of the pPCA with arbitrary 
distributions is the Variational Auto-Encoder (VAE) framework \cite{kingma2022autoencodingvariationalbayes}. 
VAEs consist of a set of variational posterior distributions that approximate the true posterior and act as a stochastic encoder. They also include a stochastic decoder that maps latent variables back to the data space, along with a prior distribution over the latent variables.
The objective is the Evidence Lower Bound (ELBO), a surrogate for the log-likelihood: a tighter ELBO yields a more accurate approximation of the data distribution, improving both sample quality and latent representations.
Unlike Flow-based Models, VAEs do not require NNs to be invertible, allowing for flexibility in the choice of architectures for both encoders and decoders. In contrast to ARMs, VAEs learn a low-dimensional latent representation of the data, providing control over the model’s information bottleneck. However, VAEs encounter several challenges, including posterior collapse 
\cite{bowman2016generatingsentencescontinuousspace}, the hole problem \cite{rezende2018tamingvaes}, difficulties in handling out-of-distribution samples \cite{Le_Lan_2021}, and a gap between the ELBO and the true log likelihood. These challenges, particularly the difficulty in handling out-of-distribution data, may compromise the applicability of VAEs in IoT domains. Nevertheless, the remarkable flexibility of this class of DGMs, exemplified by the absence of fixed NNs for both the encoder and decoder, presents a promising solution for various IoT applications \cite{jakub}.

\paragraph{\textbf{Energy-based Models}}
An Energy-based Model (EBM) \cite{inbook} is a framework in which an energy function quantifies the relationships between variables in a system. The observed inputs are represented by one set of variables, while the possible outputs or targets form another set. The energy function assigns a numerical score to each input–output configuration, with lower values indicating higher compatibility and higher values indicating lower compatibility. A Boltzmann Machine \cite{ACKLEY1985147} \cite{10.5555/104279.104290} is a specific type of EBM, where these energy scores can be transformed into probabilities, resulting in the Boltzmann (or Gibbs) distribution over possible configurations.
In practice, most energy functions yield intractable partition functions, which represent the primary computational bottleneck when training Energy-based Models. A natural extension of Boltzmann Machines are models with a deep architecture or Hierarchical Boltzmann Machines. However, training such models is even more challenging due to the complexity of the partition function, limiting their use in IoT applications \cite{jakub}. Despite the computational challenges associated with EBMs, their unconstrained nature offers a significant advantage in terms of flexibility. Specifically, the energy function is not restricted to a particular form, allowing it to be modeled by a wide variety of functions. This adaptability promotes NNs to parameterize the energy function, resulting suitable to capture highly complex relationships within the data, typical in IoT applications \cite{jakub}. Such flexibility makes Boltzmann Machines a versatile tool for representing complex systems, highlighting their potential beyond the inherent training difficulties.

\paragraph{ \textbf{Diffusion Models}}
Probabilistic models have historically faced a fundamental trade-off between two competing objectives: tractability and flexibility. Tractable models, such as Gaussian or Laplace distributions, allow for analytical evaluation and straightforward fitting to the data. However, these models are inherently limited in their ability to capture the intricate structures present in complex datasets. In contrast, flexible models are capable of representing the rich and diverse structures in arbitrary data but often sacrifice tractability, making their evaluation and optimization computationally challenging.
Diffusion Models \cite{sohldickstein2015deepunsupervisedlearningusing} 
overcome the traditional trade-off between flexibility and tractability in generative modeling. Inspired by non-equilibrium statistical physics, they first transform data into a simple prior distribution (e.g., Gaussian noise) through a forward diffusion process, then learn a reverse process that progressively removes noise to reconstruct the original data distribution. The reverse process is parameterized by NNs that predict denoising steps. By accurately approximating this process, Diffusion Models provide a flexible and tractable framework that supports efficient learning, rapid sample generation, and probability estimation, even in architectures with thousands of diffusion steps.
Moreover, Diffusion Models support conditional and posterior probability computation under the learned distribution, making them particularly versatile for a wide range of generative tasks \cite{jakub} (e.g., predictive sensor data reconstruction, anomaly-free signal synthesis in industrial settings, or synthetic telemetry generation for network traffic modeling). Hence, they excel in IoT applications, if their inherent complexities are properly handled.

\subsubsection{Implicit density}
Certain models can be trained without explicitly specifying a density function; rather, they interact indirectly with the underlying data distribution, most commonly via sampling procedures,
and fall under the second branch of the generative model taxonomy illustrated in Fig. \ref{fig:fig2}.
Within this category, some implicit models employ a Markov Chain transition operator, which must be iteratively applied multiple times to generate a sample from the model. However, Markov Chains often struggle to scale in high-dimensional spaces and incur significant computational costs. GANs were specifically designed to address these limitations. 

\paragraph{\textbf{Generative Adversarial Networks - Direct}} The Direct class includes DGMs that learn to map a simple latent distribution to complex data distributions without relying on explicit likelihood estimation. The primary example is Generative Adversarial Networks (GANs) \cite{goodfellow2014generativeadversarialnetworks}, which employ an adversarial training framework to generate high-quality samples through a competition between a generator and a discriminator.
GANs consist of two NNs, which are trained in a min-max game, such as: \textit{(i)} the generator $G$, \textit{(ii)} and the discriminator $D$. $G$ produces synthetic data samples, while the $D$ learns to distinguish between real data and the synthetic data generated by $G$. Through this adversarial process, the generator learns to create increasingly realistic samples by minimizing an adversarial loss, which measures the discriminator’s ability to correctly differentiate real and generated data. This framework allows GANs to learn complex data distributions without explicitly modeling the underlying probability density. GANs are specifically designed to overcome certain limitations associated with other types of DGMs, such as the ability to generate samples in parallel and the flexibility in designing the generator function with minimal constraints. 
For IoT applications, GANs have emerged as a powerful tool for data augmentation, anomaly detection, and privacy-preserving learning \cite{10257196, HAMOUDA2024101149, jakub}. Given that IoT systems often suffer from limited, imbalanced, or noisy datasets, GANs can generate synthetic sensor data to augment training sets, improving the robustness of machine learning models deployed in resource-constrained environments. Despite these advantages, several challenges hinder the widespread adoption of GANs in IoT. The instability of adversarial training, including mode collapse and vanishing gradients, makes training highly sensitive to hyperparameter tuning, which can be problematic in dynamic IoT environments \cite{jakub}; furthermore, GANs are computationally expensive.

\paragraph{\textbf{Markov Chain}} 
A Markov Chain is a stochastic process used to generate samples through an iterative procedure, where each new sample is drawn based on the current one according to a defined transition rule. 
However, this convergence process can be exceedingly slow, particularly in high-dimensional spaces where the efficiency of Markov Chains diminishes significantly. Moreover, a critical challenge lies in the inability to definitively determine whether the chain has reached convergence. As a result, practitioners often utilize samples prematurely, before the Markov Chain has sufficiently mixed, leading to potentially biased samples that do not accurately represent $ p_{model}(x) $. Markov Chains have broad applications across various fields, such as statistical physics, Bayesian inference, and Machine Learning, particularly in methods like MCMC. 
However, the practical challenges, including slow convergence and inefficiency in high dimensions, highlight the need for advanced techniques (e.g., Hamiltonian Monte Carlo or Variational approximations) to address these limitations and improve the scalability in modern computational problems. In the context of IoT applications, Markov Chains can be leveraged for modeling temporal dependencies in sensor data, anomaly detection, and predictive maintenance by capturing sequential patterns in time-series data \cite{s23020652}. Their ability to model probabilistic transitions makes them particularly useful for systems where state evolution follows a stochastic process. However, the practical use of Markov Chains in IoT is constrained by challenges such as computational inefficiency,  sensitivity to initial conditions, and difficulty in ensuring proper mixing \cite{app14188514}.

\begin{table}[ht!]
\footnotesize
\caption{Overview of Benefits and Limitations of DGMs in IoT Ecosystems.}
\label{tab:KOC}
\begin{center}
\raggedright
\begin{tabular}{p{2.9cm} p{2.4cm} p{4.5cm} p{4.5cm}}
\hline

\toprule
\textbf{Class} & \textbf{Model} & \textbf{IoT-related Opportunities} & \textbf{IoT-related Challenges} \\ \midrule
 
\textit{Explicit density}\newline Tractable & ARMs & Suits sequential IoT data modeling \newline Capturing long-range dependencies & Challenging parallelization due to dependencies \\ \midrule

\textit{Explicit density}\newline Tractable & Flow-based & Efficient real-time inference & Struggles with highly multi-modal data \\ \midrule

\textit{Explicit density}\newline Approximate Variational & VAEs & Efficient data compression \newline Handling of missing data & Loss of sharpness due to regularization \newline Risk of blurry reconstructions \\ \midrule

\textit{Explicit density}\newline Approximate Variational & Energy-based & Suits complex, multi-modal IoT data \newline Robust against adversarial attacks & Training challenges from unnormalized likelihood \\ \midrule

\textit{Explicit density}\newline Approximate Variational & Diffusion-based & SOTA for images data \newline High-quality data generation & Slow inference due to iterative steps \\ \midrule

\textit{Implicit density}\newline Direct & GANs & High quality data generation \newline Suits modeling complex distributions & Training instability (adversarial min-max) \\ \bottomrule

\end{tabular}
\end{center}
\end{table}

%% file: survey_sections/research_methodology.tex
\section{Research Methodology}
\label{sec:research methodology}
We conducted an initial informal search, supplemented by our personal experience: this  confirmed the existence of a substantial number of contributions to the integration of GenAI and IoT, indicating the need for a systematic review and providing also  valuable information to guide the subsequent, manual search process. Consequently, a survey of articles exploring the interaction between GenAI and IoT was conducted over a six-year period, from January 2020 to January 2026, following PRISMA guidelines, as illustrated in Fig.~\ref{fig:fig3}. This timeline aligns with the emergence and growing prominence of GenAI, as its widespread adoption and in-depth exploration began to gain traction only after 2020. Following the record screening and report selection processes, we analyzed a total of 325 studies, which were reduced to 78 after the article's review phase. The search plan undertaken is detailed in the following subsections.
\subsection{Objectives}
This article seeks to present a comprehensive overview of the current state of the art in the exploitation of GenAI for IoT Computing. To this end, a systematic survey was conducted to identify prevailing research trends, elucidate key challenges, and address the following
\textbf{Research Questions} (RQs):
\begin{itemize}
    \item (\textbf{RQ1}) What are the current scopes of GenAI for IoT applications?
    \item (\textbf{RQ2}) Which are the main GenAI models and architectures that support IoT Computing?
    \item (\textbf{RQ3}) What are the main challenges and research gaps in applying GenAI in IoT Computing? 
    \item (\textbf{RQ4}) Which are the future research directions?
\end{itemize}

\subsection{Search Strategy}
An exhaustive search of the articles regarding GenAI in IoT was performed in January 2026 by
two of the authors on the following digital libraries: ScienceDirect, Scopus, IEEEXplore,
ACM Digital Library, and Web of Science. The search focused on retrieving scientific
publications proposing solutions (i.e., models, techniques, approaches, architectures) that
aim to understand the role of GenAI in IoT.
The keyword search string was defined according to two key concepts: \textit{(i)} GenAI and \textit{(ii)} IoT. Considering such key terms, and their synonyms, the following search string was identified:
\begin{equation}
    \textit{(``Generative AI'' OR ``Generative-AI'' OR ``GenAI'' OR ``GAI'' OR ``Generative modeling')}
\end{equation}
\begin{equation}
   \textit{ AND (``Internet of Things'' OR ``IoT'' OR ``Smart devices'' OR ``Connected devices'')}
\end{equation}
 
\subsection{Eligibility criteria}
The articles were eligible for selection if they met all the following inclusion criteria: \textit{(a)}
a model or at least a definition of GenAI for IoT applications is proposed or adopted; \textit{(b)} GenAI techniques are exploited as the main element of the proposed solution; \textit{(c)} the work is an article, literature review, survey, or mapping study that specifically delves into the application of GenAI to the IoT realm.
Articles were excluded from selection if they met one of the following exclusion criteria: \textit{(a)} the terms “GenAI” and related synonyms, are contained only in the title, abstract, or keywords and are missing in the main body of the article; \textit{(b)} the concept of GenAI or one of its synonyms is either defined or used improperly; \textit{(c)} the presented GenAI techniques are not significant, or their contribution is negligible or marginal; \textit{(d)} The paper does not truly address topics, applications, or use cases related to the IoT. 

\begin{figure}[ht!]
    \centering  
    \includegraphics[width=\linewidth]{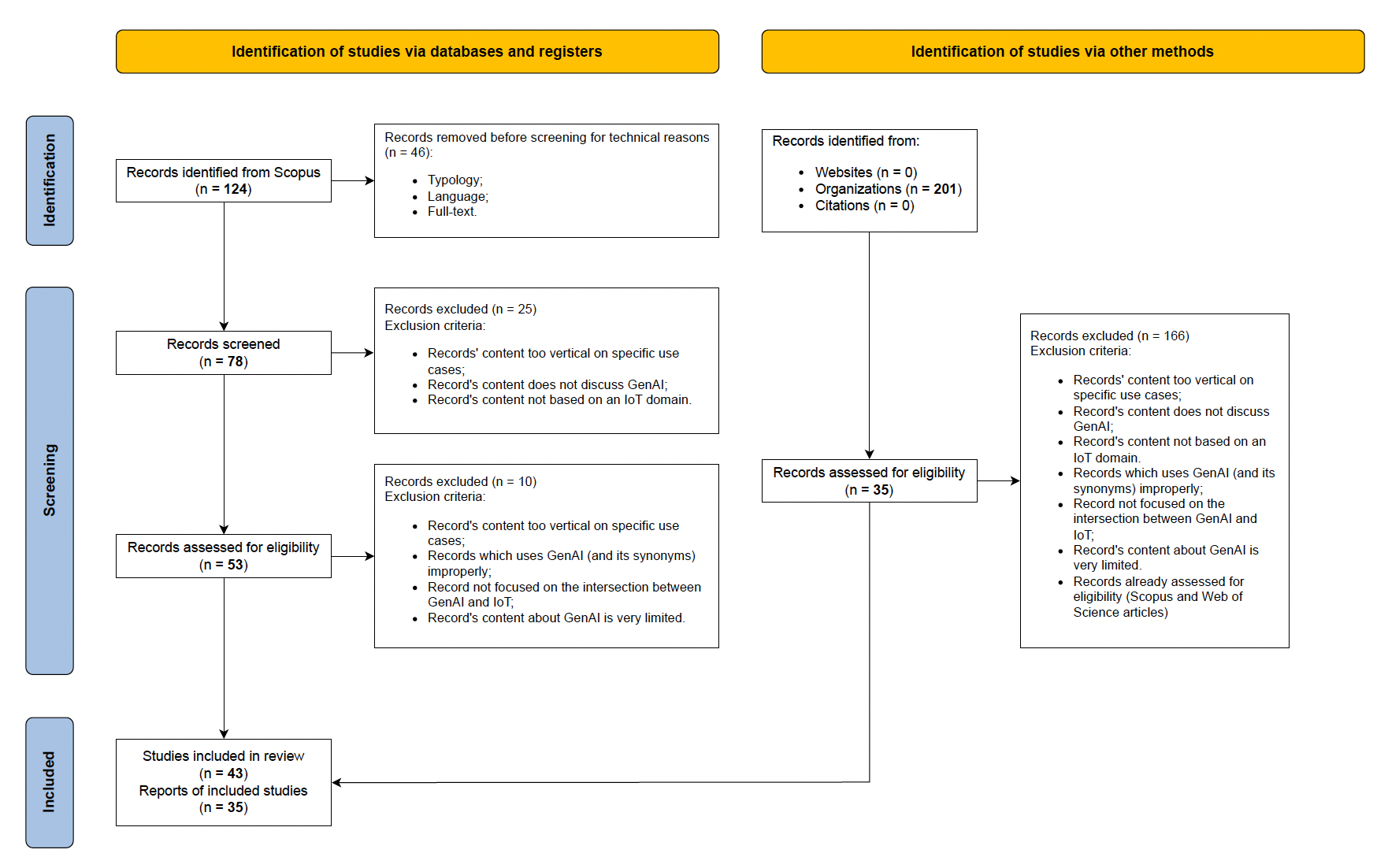}
    \caption{Flow-chart of the literature review selection process according to the PRISMA guidelines.}
    \label{fig:fig3}
\end{figure}
\subsection{Study Selection}
Fig. \ref{fig:fig3} illustrates the flow chart summarizing the approach used to select articles according to the PRISMA guidelines~\cite{PAGE2021103}. The initial search across digital libraries using the specified query retrieved \textbf{124} articles. To filter out irrelevant studies, we applied the following technical criteria: \textit{(i)} exclusion of certain publication types such as editorials, short articles, posters, theses, dissertations, brief communications, commentaries, and unpublished works; \textit{(ii)} removal of articles not partially or fully written in English; and \textit{(iii)} exclusion of papers lacking full-text availability. This initial step eliminated 46 articles, leaving 78 publications. In the first screening phase, two authors independently analyzed each article's title, abstract, and keywords based on the eligibility criteria. Any article deemed relevant by at least one author was transferred to the next phase for full-text evaluation. This phase resulted in 53 articles being selected for further review. Risk of bias was assessed during the final selection phase: all authors reviewed the full-text articles and evaluated their relevance, rigor, credibility, and quality. Manuscripts were excluded if "GenAI" was mentioned only superficially, such as in the title, abstract, or related work section, without representing a core component of the proposed solution. Inclusion decisions were made by consensus, and discussions resolved disagreements; no competing financial interests or personal relationships have influenced the study selection. Ultimately, 43 articles were selected for review. In addition, we conducted an extensive snowballing process to identify other relevant studies not captured by the initial query. This included backward (reference-based) and forward (citation-based) searches of the selected articles. We also explored gray literature, such as technical reports, Ph.D. theses, patents, and company white papers, often published by government or professional organizations. As a result, 35 additional studies were analyzed in detail. Therefore, we ended-up with 78 selected records which are analyzed in the following review.

%% file: survey_sections/literature_review.tex
\section{Literature Review}
\label{sec:literature_review}

Following the focused discussion in Section~\ref{sec:dgm} on the principles of DGMs and their theoretical applications for the IoT (see Table~\ref{tab:KOC}), we now explore how these concepts are operationalized and instantiated in the literature, according to the systematic search described in Section~\ref{sec:research methodology}. Given the large number and heterogeneity of the selected studies addressing the GenAI–IoT interplay, each article has been categorized according to the tripartite framework introduced in Section~\ref{sec:IoT}. This framework, in fact, provides a clear and intuitive conceptualization that effectively captures the multifaceted nature of IoT and serves as a solid foundation for analyzing and discussing the GenAI–IoT landscape.
In addition, this categorization not only underscores the breadth of IoT’s transformative potential but also highlights the distinct challenges and research opportunities intrinsic to each domain. By synthesizing insights across these dimensions, our analysis provides a comprehensive view of IoT's trajectory, identifying: critical trends, gaps, and emergent opportunities that inform both theoretical exploration and practical implementation strategies. Tab. \ref{tab:Comparison framework for the analyzed Internet-oriented articles 1}, Tab. \ref{tab:Comparison framework for the analyzed Things-oriented articles 1.} and Tab. \ref{tab:Comparison framework for the analyzed Semantic-oriented articles 1.}, presents a panoramic view of how recent studies are adopting GenAI methods respectively in: Internet-, Things-, and Semantic-oriented IoT perspective. Each table's row, maps an article to the four aforementioned research questions (\textit{RQ1-RQ4}), highlighting: \textit{(i)} the main focus of the article's application, \textit{(ii)} the applied DGMs according to the taxonomy introduced in Section \ref{sec:background}, \textit{(iii)} the key challenges, \textit{(iv)} and the proposed future directions. 

\subsection{Internet-oriented articles and GenAI}
The initial cluster of works (referenced in Tables~\ref{tab:Comparison framework for the analyzed Internet-oriented articles 1} and \ref{tab:Comparison framework for the analyzed Internet-oriented articles 2}) addresses several recurrent themes (\textit{RQ1}) central to Internet-oriented articles. A primary research vector involves network optimization in the context of power grids \cite{10692590}, UAV-assisted IoT networks \cite{sharif2024resourceoptimizationuavassistediot}, and 5G/6G systems \cite{10.1109/MWC.001.2300377,10750354,ferrag2023generativeaicyberthreathunting}. Parallel to this, there is a burgeoning interest in migrating GenAI to the network edge \cite{lai2023resourceefficientgenerativemobileedge,10731639}. Within this scope, GenAI facilitates diverse tasks including data augmentation, synthetic data generation \cite{ferrag2023generativeaicyberthreathunting}, multi-modal content synthesis, and semantic interpretation \cite{10596048,10773726}. Moreover, the literature demonstrates the utility of GenAI in optimizing domain-specific IoT ecosystems, notably the IoV \cite{10.1109/MWC.001.2300377}, smart cities \cite{sharif2024resourceoptimizationuavassistediot}, and IIoT \cite{wang2024joint}. Finally, a subset of the surveyed articles explores the use of GenAI to enhance the security and resilience of Internet-based frameworks \cite{10778265}. 

With respect to the main DGM and architectures (\textit{RQ2}), the literature presents a mix of explicit and implicit density approaches. ARMs \cite{10731639, ferrag2023generativeaicyberthreathunting, 10528244, 10773726} are frequently employed, appreciated for their capability to provide tractable likelihood estimates. These estimates facilitate several tasks, code generation, DGM's deploy in tactical environment, to automate 5G management and for IoV navigation systems. Hybrid solutions, incorporating  explicit and implicit density models, are also commonly adopted, as observed in \cite{ferrag2023generativeaicyberthreathunting,10596048,10.1109/MWC.001.2300377,10506539,liu2023deepgenerativemodelapplications}. These approaches are utilized for a range of objectives, application within 5G and 6G to multi-modal content generation, through mobile edge network optimization, resource optimization in UAV-assisted networks etc. Furthermore, a limited number of studies employ only Diffusion models \cite{10557650, wang2024joint} and just one only VAEs \cite{10692590}. Implicit direct density models, specifically GANs, are used only in \cite{WANG2024333}.

However, despite their differences, almost all studies acknowledge similar challenges (\textit{RQ3}), including interoperability problems, particularly within heterogeneous IoT environments where devices, protocols, and data formats must seamlessly communicate across diverse platforms. Data also gains significant prominence, as most studies emphasize the importance of high-quality and domain-specific datasets to effectively train these models within various IoT contexts, including smart cities, industrial IoT, and wireless sensor networks. Computational limitation and energy consumption (Resource-related) represent another challenge. Finally, a limited number of works address privacy and security concerns associated with the use of DGMs in Internet-oriented IoT applications \cite{ferrag2023generativeaicyberthreathunting,10596048,10.1109/MWC.001.2300377,10622951}.

Future research directions for GenAI in IoT, as synthesized from the provided tables (\textit{RQ4}), prioritize balancing computational efficiency with robust security and evolving network architectures. A major focus is on optimizing performance for constrained edge environments. Researchers propose minimizing computational overhead and enhancing device performance, employing "tiny ML" \cite{10.1109/MWC.001.2300377} and model compression techniques \cite{10731639}, and refining compression coding. Specific approaches include optimizing edge-native LLMs via fine-tuning and quantization for latency-free real-time performance, utilizing hardware-accelerated diffusion models, and developing energy-efficient hardware.
Security and privacy remain critical concerns. Solutions include developing detection-based defense strategies \cite{10596048} and secure multi-party learning frameworks \cite{10596048}, implementing secure allocation strategies, and enhancing privacy through decentralized secure training and specialized privacy-preserving methods. There is also a push to deploy blockchain \cite{liu2023deepgenerativemodelapplications} to secure against data leaks and deepfakes and to enhance "zero-trust" security architectures. Architecturally, the field is moving toward collaborative and integrated frameworks. This includes developing collaborative edge-cloud workflows, balancing edge-cloud loads \cite{10528244}, and utilizing hierarchical cloud-to-edge invocation. Research also targets integration with next-generation networks, such as 6G and evolving Open-RAN architectures \cite{10.1109/MWC.001.2300377}. Finally, ensuring reliability and advanced functionality is key. Future directions include mitigating hallucinations, integrating Knowledge Graphs \cite{10528244} for semantic analysis, improving scalability in multi-device environments, automating hyperparameter optimization and code verification tools, refining self-generating models and inference-splitting algorithms.

\begin{table}[h!]
\caption{Comparison framework for the analyzed Internet-oriented articles.}
\label{tab:Comparison framework for the analyzed Internet-oriented articles 1}
\begin{center}
\footnotesize
\raggedright
\begin{tabular}{p{0.8cm} p{3.5cm} p{2.5cm} p{3.5cm} p{4.cm}}
\hline

\toprule
\textbf{Ref.} & \textbf{RQ1} & \textbf{RQ2} & \textbf{RQ3} & \textbf{RQ4} \\ \midrule

\cite{11034683} & GenAI is used to accelerate secure digital transformation in IoT and smart cities & Implicit - Direct density (GAN) & Resource related \newline Interoperability related \newline Security related & Optimize the edge computational overhead, integration with 6G and legacy interoperability\\ \midrule 

\cite{sharif2024resourceoptimizationuavassistediot} & GenAI for UAV resource optimization, public safety, data augmentation, simulation and planning& Explicit and Implicit density & Resource related \newline Interoperability related & Integrating robotics and wireless edge computing into robust, interoperable, and ethically governed ecosystems\\ \midrule

\cite{10692590} & GenAI for optimizing power grids & Explicit - Approximate density (Variational) & Data related \newline Interoperability related & Integrate compliant multi-modal GenAI using self-generating models and data minimization technologies \\ \midrule

\cite{10557650} & GenAI to transform raw WiFi Channel State Information (CSI) into actionable insights & Explicit - Approximate density (Diffusion Models) & Data related \newline Resource related \newline Interoperability related & Optimize real-time sensing via hardware-accelerated diffusion models and  signal processing\\ \midrule

\cite{liu2023deepgenerativemodelapplications} & GenAI for optimizing network economics and routing & Explicit and Implicit density & Data related \newline Resource related \newline Interoperability related & Deploy energy-efficient edge DGMs secured by blockchain against data leaks and deepfakes \\ \midrule

\cite{10731639} & GenAI as foundation of the Internet of Senses (IoS) to transition from raw data to semantic communication & Explicit - Tractable density (ARMs) & Resource related \newline Interoperability related & Optimize edge-native LLMs via fine-tuning and quantization for latency-free, scalable real-time performance on local devices \\ \midrule

\cite{10574974} & GenAI to automate 5G management & Explicit - Tractable density (ARMs) & Data related \newline Resource related \newline Interoperability related & Accelerate network integration via LLM-automated validation and code generation for diagnostics\\ \midrule

\cite{WANG2024333} & GenAI as a proactive management layer to forecast traffic and optimize network and blockchain parameters & Implicit - Direct \newline density (GANs) & Resource related & Balance edge-cloud loads for GenAI/DRL, transitioning to physical integration, and expand blockchain predictive sharding\\ \midrule

\cite{10.1145/3638550.3641126} & LLMs as offline compilers to automate code generation for latency-constrained systems & Explicit - Tractable density (ARMs) & Data related \newline Resource related \newline Interoperability related & Enhance LLM integration, automate verification tools for code safety, explore learning mechanisms from user guidance\\ \midrule

\cite{ferrag2023generativeaicyberthreathunting} & GenAI applications within 6G-enabled IoT, for data- generation, augmentation and monitoring  & Explicit and Implicit density & Data related \newline Resource related \newline Interoperability related \newline Privacy and Security related & Enhance model scalability, develop energy-efficient hardware, decentralized secure training frameworks,  privacy-preserving strategies\\ \bottomrule

\end{tabular}
\end{center}
\end{table}

\begin{table}[h!]
\caption{Continue Tab. 3.}
\label{tab:Comparison framework for the analyzed Internet-oriented articles 2}
\begin{center}
\footnotesize
\raggedright
\begin{tabular}{p{0.8cm} p{3.5cm} p{2.5cm} p{3.5cm} p{4.cm}}
\hline

\toprule
\textbf{Ref.} & \textbf{RQ1} & \textbf{RQ2} & \textbf{RQ3} & \textbf{RQ4} \\ \midrule

\cite{10596048} & GenAI for multi-modal content generation, synthesize traffic signs for vehicular metaverses, and coordination of AI devices & Explicit and Implicit models & Resource related \newline Interoperability related \newline Privacy and Security related & Enhance efficiency and security, detection-based defense strategies, cooperative sampling, and secure multi-party learning frameworks \\ \midrule    

\cite{10528244} & GenAI to facilitate multi-modal content production and enhanced driving assistance & Explicit and Implicit density & Resource related \newline Interoperability related \newline Privacy and Security related & Collaborative edge-cloud workflows for co-inference, advanced model compression techniques \\ \midrule

\cite{10773726} & GenAI in tactical environments for code generation, intelligent services and streamlined information interpretation & Explicit - Tractable density (ARMs) & Resource related \newline Interoperability related & Enhance system resilience, improve model compression, implement secure allocation strategies, and large-scale tactical deployments \\ \midrule

\cite{wang2024joint} & GenAI to optimize decision-making in IIoT, support reliable control and  data recovery& Explicit - Approximate density (Diffusion models) & Resource related \newline Interoperability related & Improve the scalability in multi devices and dynamic environments \\ \midrule

\cite{10.1109/MWC.001.2300377} & GenAI within 5G/6G and IoV to support immersive environments, advance autonomous vehicle, zero-trust security, data augmentation& Explicit and Implicit density & Data related \newline Interoperability related \newline Security related &  Sustainability through tiny ML, support evolving Open-RAN architectures, optimize latency, enhance privacy and automate the hyperparameter optimization\\ \midrule

\cite{10622951} & GenAI in IoV for AR-aided navigation, vision-based obstacle avoidance,  in-vehicle custom experience& Explicit density (ARMs and Diffusion Models) & Data related \newline Resource related \newline Interoperability related \newline Security related & Refine inference-splitting algorithms, develop specialized collaborative mechanisms and enhance intent analysis \\ \midrule

\cite{10750354} & GenAI in 6G for customized content creation, task automation and multi-modal support & Explicit and Implicit density & Resource related \newline Interoperability related & Enhance image's quality, extend Model-Edge Generation,  personalized resource allocation, utilize DRL
\\ \midrule

\cite{lai2023resourceefficientgenerativemobileedge} & GenAI for mobile edge networks to optimize text- image and audio applications & Explicit and Implicit density & Data related \newline Resource related \newline Interoperability related & Develop privacy-preserving and defense methods, refine specialized model compression techniques\\ \midrule

\cite{10778265} & GenAI in  IoEV to support data augmentation, battery management, EV operation, grid integration, security and QoE & Explicit and Implicit density & Data related \newline Resource related & Develop adaptable and robust models, optimize vehicle-to-grid energy transactions, integrate LLM and Federated/DR Learning\\ \bottomrule 

\end{tabular}
\end{center}
\end{table}

\subsection{Things-oriented articles and GenAI}
The second cluster of works (referenced in Table \ref{tab:Comparison framework for the analyzed Things-oriented articles 1.}) addresses several recurrent themes (\textit{RQ1}) central to Things-oriented articles. A primary focus is on industrial and network optimization, utilizing GenAI in IIoT contexts for complex decision-making, combinatorial optimization, and resource scheduling \cite{ivanov2025supply}, as well as within mobile edge networks to support 6G communication and trajectory prediction \cite{lai2024ondemandquantizationgreenfederated}.
Security and system resilience also emerge as critical application areas; GenAI is applied to build self-healing systems capable of anomaly detection and automated debugging \cite{10354608}, while also enhancing public safety through active control and accident classification in the Internet of Vehicles (IoV) \cite{10517488}. Device-centric security is further explored through biometric tasks \cite{10774908}. Furthermore, specific sectors such as healthcare and smart living adopt a strong things-oriented approach. In healthcare, GenAI facilitates physiological monitoring and real-time diagnosis directly at the device level \cite{10517487}, whereas in smart home environments, it enables automated configuration and contextual understanding \cite{10632732}, or enriches AR applications for interactive education and home automation \cite{10465077} . Finally, the enhancement of user experience remains a key scope, particularly through the integration of Federated Learning to improve interactions within WSN \cite{huang2023fed}.

The majority of the reviewed works (\textit{RQ2}) employ both explicit (as ARMs, Diffusion Models, VAEs) and implicit density models (GANs) \cite{chen2025toward,huang2023fed,10354608,lai2024ondemandquantizationgreenfederated,10517488}. The use of hybrid architectures serves as a cornerstone for enhancing both systemic robustness and operational adaptability, offering an advanced solution for high-stakes application environments. Within mission-critical domains, the utilization of dual-model frameworks provides a layer of redundancy and precision. This approach is particularly instrumental in IIoT \cite{chen2025toward} for real-time process optimization, in healthcare for sensitive diagnostic support \cite{lai2024ondemandquantizationgreenfederated}, and within IoV \cite{10517488} to maintain safety and reliability in dynamic, fast-paced transit networks. Then, smaller subset of studies applies implicit models only, primarily GANs, for healthcare tasks \cite{10517487}.

Across the analyzed studies, the most frequently reported challenge (\textit{RQ3}) is related to resource constraints, including limited computational power, memory, and energy availability in IoT and edge environments \cite{chen2025toward,huang2023fed,10354608,lai2024ondemandquantizationgreenfederated,10517488}. These limitations hinder the efficient deployment of complex DGMs: thus, edge computing represents a core theme, with explicit attention to resource-efficient GenAI on-device deployment (including quantized models tailored for constrained hardware) and also scalability. Many contributions also highlight issues of interoperability, particularly in heterogeneous IoT ecosystems where seamless interaction among diverse devices and protocols is required~\cite{10632732,10774908}. In addition, the security of the device and the privacy of its data are prominent concerns, especially in applications involving personal or sensitive data, such as healthcare, smart homes, and biometric recognition~\cite{10517510,10632732,10774908}. Ensuring secure model training and inference is seen as a critical barrier to widespread GenAI adoption in such contexts. Finally, data-related challenges, such as the availability and quality of datasets for training DGMs, are also mentioned in some works, particularly where object recognition and summarization are involved~\cite{10774908,10465077}.

Looking at the future (\textit{RQ4}), a recurring direction across the literature is the optimization of GenAI models, with the aim of reducing their computational complexity, energy consumption, and improving deployment on edge devices \cite{chen2025toward,10354608,lai2024ondemandquantizationgreenfederated,10517487}. There is also attention on techniques to improve training efficiency and model validation in real-world scenarios to ensure resilience and usability \cite{chen2025toward,10774908}. This is considered essential for enabling scalable and real-time GenAI applications in constrained IoT environments \cite{chen2025toward,10517487}. Several works also anticipate future work for improving energy efficiency and scalability, particularly in edge computing scenarios where resources are limited and models must adapt to dynamic operating conditions \cite{huang2023fed,lai2024ondemandquantizationgreenfederated}. Other studies, instead, propose enhancing privacy capabilities and security, especially in domains such as healthcare and smart homes, through privacy-preserving model design, data encryption, and verification mechanisms \cite{huang2023fed, 10517488, 10632732}. Additionally, some contributions point to more domain-specific advancements, including the integration of GenAI with incentive mechanisms and blockchain for secure operations \cite{huang2023fed}, and the broader application of GenAI in self-healing systems for proactive failure resolution \cite{10354608}. 

\begin{table}[ht!]
\caption{Comparison framework for the analyzed Things-oriented articles.}
\label{tab:Comparison framework for the analyzed Things-oriented articles 1.}
\begin{center}
\footnotesize
\raggedright
\begin{tabular}{p{.8cm} p{3.5cm} p{2.5cm} p{3.5cm} p{4.cm}}
\hline

\toprule
\textbf{Ref.} & \textbf{RQ1} & \textbf{RQ2} & \textbf{RQ3} & \textbf{RQ4} \\ \midrule

\cite{chen2025toward} & GenAI in IIoT for decision making, combinatorial optimization, industrial scheduling, resource   management, offloading and energy management & Explicit density (ARMs and Diffusion models) & Resource related \newline Interoperability related & Advance diffusion-based decision-making, develop action sparsification and graph-based representations, refine training efficiency and enhance structural scalability\\ \midrule

\cite{huang2023fed} & GenAI aided by Federated learning for enhance user experience in WSN and diverse data production & Explicit and Implicit density & Resource related \newline Interoperability related \newline Security related & Design incentive mechanisms, usage of blockchain, semantic communication and AIGC-driven data augmentation, optimize edge caching\\ \midrule

\cite{10354608} & GenAI in self-healing and secure  systems for anomaly detection, automated code- generation, debugging and reporting & Explicit density (ARMs and VAEs) & Resource related \newline Interoperability related & Optimization of GenAI components, broader application domain and proactive failure resolution \\ \midrule

\cite{lai2024ondemandquantizationgreenfederated} & GenAI within mobile edge networks for various applications (e.g., trajectory prediction, education 
content generation) & Explicit and Implicit density & Resource related \newline Interoperability related & Proficient sampling for distributed diffusion, efficiency enhancement in denoising and distributed edge intelligence optimization \\ \midrule

\cite{10517488} & GenAI in IoV for data augmentation, accident classification, active safety control and real time monitoring & Explicit and Implicit density & Data related \newline Resource related \newline Interoperability related \newline Privacy and Security related & Device compatibility and interoperability, data encryption and privacy, enhancing model interpretability and communication framework \\ \midrule

\cite{10465077} &  GenAI in AR apps for home automation, interactive and education experiences, multimedia content retrieval, etc.& Not explicitly mentioned & Data related \newline Resource related & Refining computer vision, enhancing social interaction and remote, integrating AI chatbots and expanding smart home connectivity\\ \midrule

\cite{10774908} & GenAI for fingerprint reconstruction, weapon and threat detection and suspicious activity recognition & Explicit and Implicit density & Data related \newline Resource related \newline Interoperability related \newline Privacy related & Real-world testing and validation, usability improvements, resilience and flexibility and synthetic data refinement \\ \midrule

\cite{10517487} & GenAI in healthcare for physiological monitoring, data augmentation, real-time diagnosis& Implicit - Direct density (GANs) & Resource related & Ecosystem development, optimize latency, reduce edge computing power and enhance battery life \\ \midrule

\cite{10632732} & GenAI for Smart Home for automated configuration and contextual understanding  & Explicit - Tractable density (ARMs) & Interoperability related  \newline Privacy and Security related & Standardized interfaces, verification and safety mechanism, local execution and feedback loops\\ \midrule
\cite{10517510} & GenAI for domains specific applications, optimize user interfaces and infrastructure, data generation and processing& Explicit and Implicit density & Data related \newline Resource related \newline Privacy and Security related & Developing resilient models, enhance efficiency and sustainability, enhance security and establish universal protocols for standardization \\ \bottomrule

\end{tabular}
\end{center}
\end{table}

\begin{table}[h!]
\caption{Comparison framework for the analyzed Semantic-oriented articles.}
\label{tab:Comparison framework for the analyzed Semantic-oriented articles 1.}
\begin{center}
\footnotesize
\raggedright
\begin{tabular}{p{.8cm} p{3.5cm} p{2.5cm} p{3.5cm} p{4.cm}}
\hline

\toprule
\textbf{Ref.} & \textbf{RQ1} & \textbf{RQ2} & \textbf{RQ3} & \textbf{RQ4} \\ \midrule

\cite{ivanov2025supply} & GenAI to integrate "point solutions", support decision making, automatic modeling and risk assesment & Explict - Tractable density (ARMs) & Data related \newline Resource related \newline Interoperability related \newline Security related & Extending GenAI to the downstream supply chain, enhance interoperability, optimize human-AI collaboration, support decision making\\ \midrule

\cite{10609561} & GenAI for device defect detection, network intrusion detection, few-shot learning,  simulation and data augmentation &  Explicit and Implicit density & Data related \newline Resource related \newline Interoperability related \newline Privacy and Security related & Enhance scalability, improving efficiency, address security and privacy issue \\ \midrule

\cite{10517500} & GenAI for data augmentation and generation, intelligent orchestration, human-machine interaction, decision support and security monitoring & Not specifically mentioned & Data related \newline Resource related \newline Privacy and Security related & Address security issue in multi-agent systems and developing scalable security frameworks \\ \midrule

\cite{info15110740} & GenAI for data augmentation, to enhance threat detection and context aware simulation & Explicit and Implicit density& Data related \newline Resource related & Enhancing security capabilities by extending this framework to accommodate a broader range of attack\\ \midrule

\cite{10517484} & GenAI for traffic modeling, prediction, management and optimization, anomaly detection, wireless channel prediction& Explicit and Implicit density & Data related \newline Resource related \newline Interoperability related \newline Privacy and Security related & Developing resilient models, improve privacy techniques and enhance model compression techniques \\ \midrule

\cite{li2023fillingmissingexploringgenerative} & GenAI for data augmentation, content creation and to enable low-latency services & Explicit and Implicit density & Data related \newline Resource related \newline Privacy related & Multi-server content generation \\ \midrule

\cite{10720312} & GenAI for cyber-security scopes, anomaly detection and Federated learning integration & Explicit - Tractable density (ARMs) & Data related \newline Resource related & Hyperparameter optimization, enhance data pre-processing and handling imbalanced data \\ \midrule

\cite{ferrag2024revolutionizingcyberthreatdetection} & GenAI for cyber-threat detection, malware detection, log and system anomaly detection & Explicit - Tractable density (ARMs) & Data related \newline Resource related \newline Privacy and Security & Expanding threat coverage, coninously updating models and autonomous migration \\ \midrule

\cite{10778718} & GenAI for synthetic data generation and personalized learning & Explicit - Tractable density (ARMs) & Resource related & Models optimization \\ \midrule

\cite{10778271} & GenAI in IoV to enhance connectivity and reliability, data processing,  augmentation, anomaly detection& Implicit - Direct density (GANs) & Resource related \newline Interoperability related \newline Privacy and Security related & Real-time decision making in large-scale networks, integration of RL, security optimization, and improving scalability\\ \bottomrule

\end{tabular}
\end{center}
\end{table}

\begin{table}[h!]
\caption{Continue Tab. 6.}
\label{tab:Comparison framework for the analyzed Semantic-oriented articles 2.}
\begin{center}
\footnotesize
\raggedright
\begin{tabular}{p{.8cm} p{3.5cm} p{2.5cm} p{3.5cm} p{4.cm}}
\hline

\toprule
\textbf{Ref.} & \textbf{RQ1} & \textbf{RQ2} & \textbf{RQ3} & \textbf{RQ4} \\ \midrule
\cite{HAMOUDA2024101149} & GenAI in IIoT for data augmentation, privacy and security protection & Implicit - Direct density (GANs) & Data related \newline Resource related \newline Privacy and Security related & Supervised learning to further enhance model's capabilities, improve security mechanism\\ \midrule

\cite{10612836} & GenAI in UAV-based mission critical networks for trajectory optimization, channel estimation and modeling& Explicit and Implicit density & Data related \newline Resource related \newline Interoperability related \newline Privacy and Security related & Compressing GenAI models, integration of GenAI with O-RAN, develop real-time semantic communications and improve security\\ \midrule

\cite{10517486} & GenAI for voice assistant, smart surveillance, healthcare monitoring, traffic monitoring& Explicit and Implicit density& Resource related \newline Interoperability related \newline Privacy and Security related & Developing lightweight models, improve security, investigate GenAI role in real-time tasks\\ \midrule

\cite{10257196} & GenAI for cyber-threat and intrusion detection  & Implicit - Direct density (GANs) & Data related \newline Privacy and Security related & Not explicitly mentioned\\ \midrule

\cite{10729865} & LLMs for Smart Home environments & Explicit - Tractable density (ARMs) & Data related \newline Resource related \newline Interoperability related & Model's edge deployment for privacy reasons, latency optimization and multi-modal approach \\ \midrule

\cite{en17081935} & GenAI for data analytics, decision support and efficiency, anomaly detection and forecasting & Explicit - Tractable density (ARMs) & Data related \newline Resource related \newline Interoperability related \newline Privacy and Security related & Advanced model integration, real-time data processing, enhanced personalization and multi-modal interaction\\ \midrule

\cite{das2023changemanagementusinggenerative} & GenAI for data generation, digital twin development and system integrity monitoring & Explicit and Implicit density & Data related \newline Resource related & Advanced model integration \\ \midrule

\cite{10463265} & GenAI for Digital Twin integration, predictive maintenance and real-time monitoring  & Explicit and Implicit density & Data related \newline Resource related \newline Interoperability related & Security optimization and human in the loop\\ \midrule

\cite{10578567} & GenAI for data interpretation and pattern identification & Explicit - Tractable density (ARMs) & Data related & Not explicitly mentioned \\ \midrule

\cite{10716549} & GenAI for traffic analysis and performance evaluation, security and network management& Explicit and Implicit density & Data related \newline Resource related \newline Interoperability related & Model's optimization and improved evaluation metrics  \\ \midrule

\cite{9326416} & GenAI for data compression & Explicit - Approximate density (VAEs) & Data related \newline Resource related & Model optimization\\ \midrule
\cite{10578894} & GenAI for personalized learning & Explicit - Tractable density (ARMs) & Interoperability related & Educational integration and more extensive validation \\ \midrule

\cite{10577097} & GenAI for data generation, simulation and  optimization & Explicit and Implicit density & Resource related \newline Interoperability related & Model optimization \\ \midrule
\cite{10506539} & GenAI for navigation optimization, traffic prediction, evaluation and map generation & Explicit and Implicit density & Resource related \newline Interoperability related \newline Privacy and Security related & Develop Real-time models, enhance privacy and security and investigate the collaboration edge-cloud \\ \midrule
\cite{10592370} & GenAI for network optimization and resource allocation  & Explicit - Tractable density (ARMs) & Resource related \newline Interoperability related & Model optimization \\ \bottomrule 
\end{tabular}
\end{center}
\end{table}

\begin{table}[h!]
\caption{Continue Tab. 7.}
\label{tab:Comparison framework for the analyzed Semantic-oriented articles 3.}
\begin{center}
\footnotesize
\raggedright
\begin{tabular}{p{.8cm} p{3.5cm} p{2.5cm} p{3.5cm} p{4.cm}}
\hline

\toprule
\textbf{Ref.} & \textbf{RQ1} & \textbf{RQ2} & \textbf{RQ3} & \textbf{RQ4} \\ \midrule

\cite{10463583} & GenAI for data generation, safety hazards mitigation and enhancing monitor techniques & Explicit - Tractable density (Diffusion Models) & Data related & Evaluation with real-world data\\ \midrule

\cite{10261616} & GenAI in Agriculture 4.0 for data augmentation/integration, scenario simulation/generation& Explicit and Implicit density & Data related \newline Resource related \newline Privacy and Security related & Further improve the duo GenAI - Digital Twin and overcoming the implementation barriers \\ \midrule

\cite{9639241} & GenAI for oversampling imbalanced data & Explicit and Implicit density & Data related \newline Resource related & Model optimization \\ \midrule

\cite{du2023yolobasedsemanticcommunicationgenerative} & GenAI for resource optimization, semantic communication and data trasmission & Explicit and Implicit density & Resource related & Model optimization \\ \midrule

\cite{GILL2023262} & GenAI for Smart home, autonomous vehicles, IIoT, Healthcare, Robotics and Smart city   & Explicit - Tractable density (ARMs) & Data related \newline Resource related \newline Interoperability related \newline Privacy and Security related & Enhance Human-AI communication, improve energy efficiency, address cyber-security risks, deepening IoT integration\\ \midrule

\cite{10566961} & GenAI for data generation, privacy preserving data sharing and empowering Digital Twins & Explicit and Implicit density & Data related \newline Resource related \newline Privacy and Security related & Real-World deployment, standardization for benchmarking\\ \midrule

\cite{10517485} & GenAI for data generation and to handle uncertainty in IoT data & Explicit and Implicit density & Data related \newline Resource related \newline Interoperability related & Model optimization\\ \midrule

\cite{10520929} & GenAI for adversarial defense, data storage and retrieval and resource allocation & Explicit - Approximate density (Diffusion Models) & Resource related \newline Interoperability related \newline Privacy and Security related & Model optimization, enhance energy efficiency and improve adaptability\\ \midrule

\cite{10467761} & GenAI for security, resource allocation and image retrieval & Explicit - Approximate density (Diffusion) & Resource related \newline Privacy and Security related & Improving efficiency, security and optimize model's generalizability\\ \midrule

\cite{wang2024internetthingseragenerative} & GenAI for Mobile networks, Autonomous vehicles, Cyber-security, Robotics and Healthcare & Explicit - Tractable density (ARMs) & Data related \newline Resource related \newline Interoperability related \newline Privacy and Security related & Building IoT-specific DGMs, advanced model compression techniques, improve on-device fine-tuning \\ \midrule
\cite{10517482} & GenAI for traffic and workload prediction and system optimization & Implicit - Direct density (GANs) & Data related \newline Resource related \newline Interoperability related & Model optimization \\ \midrule
\cite{10718527} & GenAI for data generation and predictive maintenance& Explicit and Implicit density & Data related \newline Resource related & Model optimization and improve privacy mechanism\\ \midrule
\cite{ai5030055} & GenAI for data generation & Explicit - Tractable density (ARMs) & Data related \newline Privacy and security related & Model optimization \\ \midrule
\cite{10759478} & GenAI for real-time management, driver assistance and simulation & Implicit - Direct density (GANs) & Data related \newline Interoperability related \newline Privacy related & Improve scalability and model optimization \\ 
\bottomrule
\end{tabular}
\end{center}
\end{table}

\subsection{Semantic-oriented articles}
In this third cluster of studies (referenced in Table \ref{tab:Comparison framework for the analyzed Semantic-oriented articles 1.}, Table \ref{tab:Comparison framework for the analyzed Semantic-oriented articles 2.} and Table \ref{tab:Comparison framework for the analyzed Semantic-oriented articles 3.}), a predominant area of application \textit{(RQ1)} is the security and management. Several studies focus on GenAI for cyber-threat detection, intrusion detection and malware identification \cite{ferrag2024revolutionizingcyberthreatdetection,10720312}. A second fundamental pillar is represented by data management and simulation. The literature highlights a massive use of GenAI for data augmentation and synthetic data generation, which are essential for IoT applications \cite{10517500, info15110740, li2023fillingmissingexploringgenerative,10778718}. Finally GenAI is transforming smart environments and services centered on the user. In smart home and healthcare sectors LLMs are used for voice assistant and smart surveillance \cite{10517486, 10729865}.

Regarding the models and techniques adopted (\textit{RQ2}), the literature presents a diverse landscape of DGMs. A significant portion of the analyzed works relies on implicit density estimation leveraging the GANs \cite{10778271, HAMOUDA2024101149, 10257196} for tasks as cyber-threat detection, traffic prediction and real-time management. 
In contrast, explicit density models are widely used, particularly those classified as having tractable density such as ARMs. These are frequently applied in resource constrained environments like Smart homes and mobile network due to their extensive research in model compression techniques and ability to analyze sequential data \cite{ivanov2025supply, 10720312, ferrag2024revolutionizingcyberthreatdetection}. Then another emerging category is represented by Diffusion models, which are increasingly applied for complex decision-making, adversarial defence and safety hazard mitigation \cite{10520929,10467761,10463583}. Additionally, some works utilize Approximate Density methods, specifically Variational Autoencoders (VAEs), often for data compression or self-healing systems \cite{9326416}. Finally, a substantial number of articles adopt a Hybrid approach, combining Explicit and Implicit density techniques. This trend suggests a need to leverage the strengths of multiple architectures to address complex challenges \cite{10716549,10463265}.

Addressing the challenges and key aspects identified in the literature (\textit{RQ3}), the analysis reveals that resource constraints represent the most pervasive hurdle. A vast majority of studies highlight the difficulty of deploying GenAI models on edge devices due to limited computational power, memory, and energy supply. This challenge is central to research in IIoT, healthcare, and network optimization \cite{ivanov2025supply,10609561,10517500,info15110740}.
Closely linked to physical resources are data-related challenges. The literature emphasizes issues regarding data availability, quality, and management. Specific concerns include handling unbalanced datasets, data compression, and the interpretation of complex patterns, which are critical for effective training and inference \cite{10517484,li2023fillingmissingexploringgenerative,10720312}. Furthermore, privacy and security concerns are paramount, particularly in sensitive domains such as healthcare, smart homes, and critical infrastructure. The challenges range from protecting data against adversarial attacks and unauthorized access to ensuring privacy during data generation and sharing. Several works focus specifically on these aspects, developing solutions for intrusion detection, malware defense, and secure data handling \cite{ferrag2024revolutionizingcyberthreatdetection,10778271,HAMOUDA2024101149}. Interoperability also remains a critical requirement, especially in heterogeneous environments where different devices and systems must communicate seamlessly. This is highlighted in studies focusing on personalized learning, smart home automation, and general simulation tasks \cite{10612836,10517486}. Finally, it is notable that a significant portion of the analyzed articles adopts a holistic view, addressing multiple challenges simultaneously.

Looking at the future research directions (\textit{RQ4}), the literature overwhelmingly points towards the need for model optimization and efficiency. A vast number of studies prioritize reducing the computational burden of GenAI models to make them suitable for resource-constrained IoT and edge environments. This includes general model optimization \cite{10778718} as well as specific efforts to reduce energy consumption, latency, and battery usage \cite{10517510,10517510,10354608}. Several works explicitly propose the development of lightweight models and advanced compression techniques to facilitate on-device deployment \cite{10517484, 9326416}. A second critical area of future work is Security, Privacy, and Trust. Researchers emphasize the necessity of enhancing privacy-preserving mechanisms, often through encryption, blockchain integration, or secure model design \cite{10609561,10517500,info15110740}. 
Scalability and Interoperability are also major concerns for the next generation of GenAI-IoT systems. Future research aims to improve structural scalability for large-scale industrial applications and autonomous networks \cite{10609561,10778271}. This includes developing standardized interfaces and ensuring device compatibility, as well as investigating collaboration models between edge and cloud architectures.
Finally, there is a strong push for testing models with real-world data and establishing standardized benchmarking \cite{10463583, 10566961}.
\subsection{Cross-comparison}
To provide a stronger comparative synthesis beyond narrative reporting, Table \ref{tab:qualitative_cross_analysis} presents a cross-comparison of the reviewed literature. For the subset of studies that provided sufficient elements to conduct this analysis, such as the availability of open-source code, public datasets, or comprehensive experimental setups, we assessed their relative strengths, weaknesses, and overall maturity. However, given the significant heterogeneity among the surveyed papers regarding their specific methodological techniques and individual research objectives, a standardized quantitative evaluation proved infeasible. Hence, we performed a qualitative cross-comparison pivoted around key objective criteria: \textbf{Latency}, \textbf{Scalability}, \textbf{Energy Efficiency}, \textbf{Performance}, \textbf{Reproducibility}, and \textbf{Deployment Readiness}.

\begin{table*}[hbt!]
\small
\centering
\caption{Qualitative cross-comparison against key objective criteria.}
\label{tab:qualitative_cross_analysis}
\renewcommand{\arraystretch}{1.3}
\resizebox{\textwidth}{!}{
\begin{tabular}{p{0.05\linewidth} p{0.18\linewidth} p{0.18\linewidth} p{0.18\linewidth} p{0.18\linewidth} p{0.18\linewidth} p{0.18\linewidth}}
\toprule
\textbf{Article} & \textbf{Latency} & \textbf{Scalability} & \textbf{Energy Efficiency} & \textbf{Performance} & \textbf{Reproducibility} & \textbf{Readiness} \\ \bottomrule

\multicolumn{6}{l}{\textit{\textbf{Internet-oriented articles}}} \\ \toprule

\textbf{\cite{ferrag2023generativeaicyberthreathunting}} & \textbf{High}: DGMs inference on edge/cloud nodes  & \textbf{Medium}: Attempts to load-balance& \textbf{Low}: Heavy energy and resource consumption & \textbf{High}: Overall accuracy >95\%
& \textbf{Low}: No available dataset and code & \textbf{Low}: Validated in a theoretical setting\\ \midrule

\textbf{\cite{lai2023resourceefficientgenerativemobileedge}} & \textbf{Medium}: Model partitioning to load distribution & \textbf{High}: Architecture’s core focus & \textbf{High}: Architecture’s core focus & \textbf{High}: Outperform baselines 
& \textbf{Medium}: No available code & \textbf{Low}: Validated in a theoretical setting \\ \midrule

\textbf{\cite{10750354}} & \textbf{Medium}: Overhead to the local side & \textbf{High}: Reducing the load alleviates network congestion & \textbf{Medium}: Architecture’s core focus & \textbf{High}: Reduced communication, better robustness and results quality 
& \textbf{Medium}: No available code & \textbf{Low}: Validated in a theoretical setting \\ \midrule

\textbf{\cite{10731639}} & \textbf{Medium}: Due to multiple LLMs & \textbf{High}: Due to bandwidth reduction & \textbf{Medium}: Energy burden on cloud servers & \textbf{High}: Outperform baselines 
& \textbf{High}: Methodologically replicable framework & \textbf{Medium}: Validated on a prototype \\ \midrule

\textbf{\cite{10.1109/MWC.001.2300377}} & \textbf{Medium}: Model partitioning to distribute the load & \textbf{Medium}: Potential processing bottleneck for large connected fleets & \textbf{Low}: Cloud constantly run multiple energy-intensive tasks & \textbf{Medium}: Reduction of model inversion and model poisoning attacks 
& \textbf{Medium}: Only partially accessible & \textbf{Low}: Conducted as a simulated experiment \\ \bottomrule 

\multicolumn{6}{l}{\textit{\textbf{Thing-oriented articles}}} \\ \toprule

\textbf{\cite{ivanov2025supply}} & \textbf{Medium}: Heavy inference on cloud & \textbf{High} Architecture's core focus & \textbf{Low}: Massive data ingestion and LLMs orchestration & \textbf{High}: Substantial qualitative operational improvements & \textbf{Low}: No available dataset and code & \textbf{High}: Implemented and tested in a large-scale operational setting  \\ \midrule 

\textbf{\cite{lai2024ondemandquantizationgreenfederated}} & \textbf{Medium}: Diffusion models require long computation  & \textbf{Medium}: Aggregation on a central server could create bottlenecks & \textbf{High}: Architecture's core feature & \textbf{High}: Reduced energy and model size vs. baselines & \textbf{Medium}: No available code  & \textbf{Low}: Validated exclusively through numerical simulations \\ \midrule 

\textbf{\cite{10354608}} & \textbf{Low}: High overhead & \textbf{Medium}: Individual request to LLM & \textbf{Low}: LLM and VAE processing & \textbf{High}: Strong code generation capabilities & \textbf{Medium}: Core logic tied to proprietary models & \textbf{Medium}: Working prototype tested in a sandbox environment  \\ \midrule 

\textbf{\cite{10517488}} & \textbf{High}: Decentralized AI inference on edge & \textbf{High}: Distributed architecture & \textbf{Medium}: Edge devices forced to heavy processing & \textbf{Medium}: Enhance timeliness, accuracy and stability & \textbf{Low}: No available code & \textbf{Low}: Purely theoretical/exploratory study\\ \midrule 

\textbf{\cite{10774908}} & \textbf{Medium}: Due to the offloading of multimedia data & \textbf{Medium}: HW requirements are hindering its widespread adoption & \textbf{Low}: Multimedia data processing & \textbf{Medium}: Lacks quantitative performance evaluations 
& \textbf{Medium}: Purely theoretical/exploratory study & \textbf{Low}: It is merely an architectural design  \\ \bottomrule 

\multicolumn{6}{l}{\textit{\textbf{Semantic-oriented articles}}} \\ \toprule

\textbf{\cite{li2023fillingmissingexploringgenerative}} & \textbf{Medium}: Overhead caused by data aggregation on the server & \textbf{High}: Distributed architecture & \textbf{High}: Architecture's core feature & \textbf{High}: Reduced energy consumption vs. baselines & \textbf{Medium}: No available code & \textbf{Medium}: Conducted primarily through software simulations  \\ \midrule 

\textbf{\cite{10517486}} & \textbf{Low}: Diffusion models requires several denoise step & \textbf{High}: Architecture's core feature & \textbf{Medium}: Due to Diffusion Models & \textbf{High}: Improve accuracy vs. baselines 
& \textbf{Low}: No available code and dataset & \textbf{Low}: The framework is tested in conceptual and mathematical settings  \\ \midrule 

\textbf{\cite{10517500}} & \textbf{Medium}: AI inspection introduce delays & \textbf{Medium}: Significant challenges in coordination and encryption & \textbf{Low}: Due to AI algorithm involvement & \textbf{Low}: Comprehensive analysis, no proposed model  & \textbf{Low}: Comprehensive analysis no proposed model & \textbf{Low}: High-level design discussion  \\ \midrule 

\textbf{\cite{10729865}} & \textbf{Low}: Chain-of-Thought plus cloud LLM APIs adds noticeable  & \textbf{High}: Architecture's core feature & \textbf{Low}: Energy-intensive models & \textbf{High}: Outperforming baselines & \textbf{High}: Available code & \textbf{High}: The framework has been integrated and validated in practice  \\ \midrule 

\textbf{\cite{ferrag2024revolutionizingcyberthreatdetection}} & \textbf{High}: Architecture's core feature & \textbf{High}: Low footprint & \textbf{High}:Low energy consumption due to low footprint & \textbf{High}: Outperforming baselines & \textbf{Medium}: No available code & \textbf{Medium}: Tested on a realistic testbed, with real hardware  \\ \bottomrule 

\end{tabular}
}
\end{table*}
The comparative analysis suggests that the field is currently in a transition phase from proof-of-concept architectures toward deployable AI-enabled IoT systems. Across all categories of studies, performance emerges as the most mature dimension, with the majority of approaches achieving high effectiveness regardless of their architectural orientation. Similarly, scalability is generally well addressed through distributed processing, model partitioning, and decentralized intelligence mechanisms, indicating that large-scale operation has become a primary design objective. In contrast, energy efficiency remains a significant challenge, as the computational demands of LLMs, diffusion models, cloud offloading, and continuous data processing often result in low or moderate efficiency levels. Reproducibility represents the weakest aspect of the surveyed literature, with many works lacking publicly available code, datasets, or experimental artifacts, thereby limiting independent validation and comparative assessment. Deployment readiness also remains relatively low, as most solutions are evaluated through simulations, conceptual studies, or prototype implementations. Notably, all the approaches exhibit comparable performance levels, suggesting that the main differences among them lie in their trade-offs between scalability, energy consumption, reproducibility, and practical maturity rather than in raw effectiveness. Finally, the few studies that achieve both high reproducibility and high deployment readiness are typically those providing open-source implementations and undergoing practical validation, highlighting the key role of open science  practices in boosting the transition from experimental to operational GenAI-enabled IoT systems.

%% file: survey_sections/practical_challenges_of_generative_ai_for_iot_systems.tex
\section{Bridging the Gap: Solutions for the Practical Challenges of Generative AI in IoT Computing}
\label{sec:challenges}

As outcome of the discussion of the previous Section \ref{sec:literature_review}, it can be concluded that the development of GenAI-aided IoT systems presents both substantial opportunities and considerable challenges. Following the tripartite framework classification, a summary of main, high-priority challenges for each vision include:\\
\emph{Challenges in Thing-oriented vision.} The integration of GenAI into IoT ecosystems introduces a fundamental conflict between model complexity and the capabilities of edge hardware and IoT devices, making it challenging to meet the computational, memory, and energy demands required by GenAI models and the peculiar  time constraints required by industrial applications. Hence, to improve efficiency while maintaining DGM accuracy is crucial~\cite{Cai_2022}.\\ 
\emph{Challenges in Network-oriented vision.} Ensuring scalability, interoperability across heterogeneous platforms, and efficient network support for applications that are intrinsically cloud-centric are well-known IoT challenges. However, these are further exacerbated by  GenAI workloads, that significantly increase network traffic and place stringent demands on communication reliability, bandwidth availability and end-to-end latency \cite{liu2023deepgenerativemodelapplications}.\\
\emph{ Challenges in Semantic-oriented vision.} The quality and quantity of the data used for GenAI  training or fine-tuning is not a negligible aspect, but, at the same time, sensitive domains like Smart Healthcare involve privacy-sensitive data, limiting the feasibility of centralized data collection and training. Moreover, unlike general-purpose domains that benefit from massive public repositories, the IoT landscape suffers from severe data scarcity and fragmentation, since real-world sensor data are often proprietary, sensitive, or, in the case of fault detection, statistically rare, making it difficult to construct balanced datasets for supervised learning. These limitations directly impact the ability of GenAI models to learn robust semantic representations of physical-world processes \cite{10517500}.

\begin{figure}[ht!]
    \centering  \includegraphics[width=.65\linewidth]{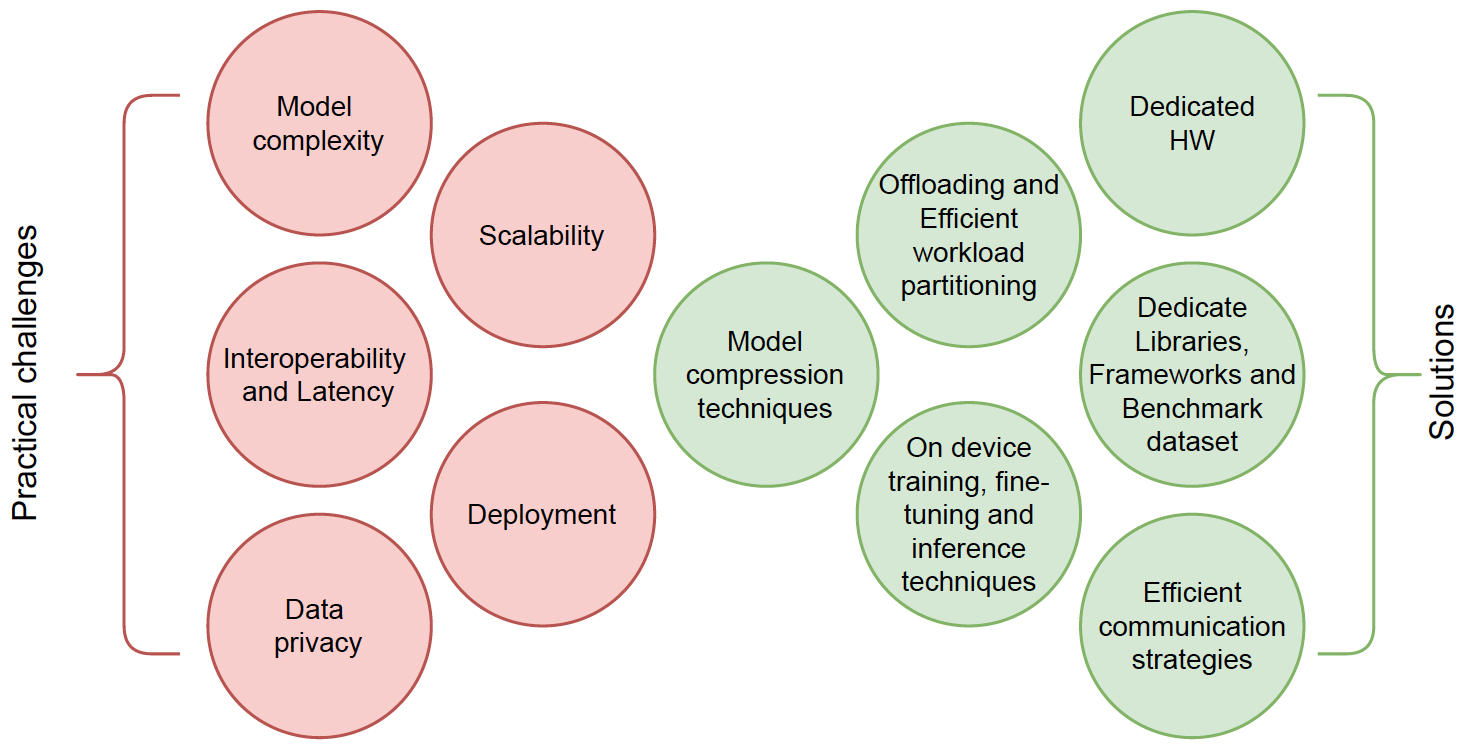}
    \caption{Solutions for the Practical challenges of GenAI for IoT Computing.}
    \label{fig:fig9}
\end{figure}

Furthermore, deploying GenAI-aided IoT systems presents itself as a significant, \textit{cross-vision challenge}: specific implementation, benchmarking, and regulation aspects still remain largely unexplored, and relevant topics pertaining, for example, ethical concerns,
explainability, trust, and security, scarcely addressed. 
Therefore, moving beyond the study-specific findings of RQ4, this section examines \textbf{transversal solutions} to overcome the practical challenges, thus making DGMs viable for IoT Computing (see Fig. \ref{fig:fig9}).

\subsection{Model compression techniques}
DGMs are sophisticated architectures characterized by billions of parameters, designed to adhere to the scaling law. This principle posits that achieving higher levels of accuracy and performance requires increasingly larger model architectures. To speed up the time required for inference, reduce the number of computational resources needed, and consequently reduce the environmental impact of such models, a number of model compression techniques have been proposed. 
A summary is then provided by Tab. \ref{tab:solutions}.
\paragraph{\textbf{Pruning}}
Parameter pruning encompasses several approaches that differ in granularity and hardware implications. Fine-grained  pruning \cite{han2015learningweightsconnectionsefficient} removes individual elements, such as weights, from data structures like tensors. Owing to its high level of granularity, this approach enables selective pruning and can achieve substantial compression ratios while largely preserving model accuracy. Pattern-based pruning \cite{niu2020patdnn}represents a specialized variant of fine-grained pruning, in which predefined sparsity patterns are exploited to facilitate hardware acceleration through compiler-level optimizations. In contrast, coarse-grained pruning operates at a higher level by removing entire tensor blocks, thereby improving hardware efficiency and enabling direct acceleration on GPUs when standard deep learning libraries are used. However, this reduced granularity typically comes at the cost of lower accuracy compared to fine-grained pruning.
  
Conventional pruning schemes are more compatible with hardware architectures, leading to reduced energy consumption and enhanced inference acceleration making them suitable in adapting DGMs for IoT resource-constrained environments. In contrast, irregular pruning schemes tend to better preserve model accuracy at equivalent compression rates, but resulting in a higher hardware resource consumption. However, with advances in specialized hardware accelerators \cite{chen2019eyerissv2flexibleaccelerator} and compiler-based optimization techniques \cite{li2023fillingmissingexploringgenerative}, significant acceleration can also be achieved for irregular pruning methods, also making such techniques suitable for IoT applications. Once the pruning granularity has been determined, the selection of weights to be pruned is critical to the performance of the pruned model. While numerous methodologies have been proposed, they all adhere to the fundamental principle of removing less significant weights based on predefined criteria. The most straightforward heuristic relies on magnitude, where weights with larger absolute values $|W|$ are considered more critical \cite{han2017deeppyramidalresidualnetworks}. 
Alternative criteria include second-order derivatives \cite{NIPS1992_303ed4c6}, loss approximation via Taylor expansion \cite{molchanov2017pruningconvolutionalneuralnetworks}, and output sensitivity \cite{963775}. Directly removing weights from DGMs, will affect the accuracy of the models. To mitigate the risk of improperly pruning essential weights, \textit{Dynamic pruning} \cite{guo2016dynamicnetworksurgeryefficient} (integrating connection splicing as a continuous network maintenance task) and \textit{Runtime pruning} \cite{NIPS2017_a51fb975} (adjusting pruning ratios based on input samples) have been proposed. Both techniques, albeit through different mechanism, transform pruning from a static procedure into a dynamic one. Although pruning-supported techniques have been employed in IoT applications for compressing Deep Neural Networks (DNNs) \cite{9691274, 9442600}, their application to DGMs in IoT environments remains unexplored. To the best of our knowledge, no existing studies have investigated the use of pruning techniques for DGMs in this context.

\paragraph{\textbf{Low Rank Approximation (LoRA)}}
LoRA \cite{hu2021loralowrankadaptationlarge} 
utilizes the concept of approximating the original weight matrix by decomposing it into two or more smaller matrices with lower dimensions, then reducing the size and complexity of DGMs. To mitigate potential information loss, various techniques have been developed, broadly classified into two categories: 
   \textit{Training-required methods}, which  require fine-tuning the entire model either during or after the application of LoRA to restore or improve overall performance \cite{ben-noach-goldberg-2020-compressing}; and  \textit{Training-free methods}, which  prioritize selecting the least
    significant matrices for decomposition, thereby reducing complexity without requiring
    additional training \cite{hua2022numericaloptimizationsweightedlowrank}.

A widely applied technique is Truncated Singular Value Decomposition (SVD) \cite{doi:10.1137/1.9781421407944}. The LoRA approximation does not require specialized hardware for implementation and execution, making it a highly suitable solution for applications in IoT domains.
From the analyzed literature, it emerges that one study employs LoRa technique \cite{huang2023fed}. However, recent studies have yet to explore the compression of DGMs using LoRA for deployment on IoT devices.

\paragraph{\textbf{Quantization}}
Network quantization 
reduces memory requirements and computational costs by decreasing the number of bits needed to represent weights in a DNNs. 
Early approaches \cite{gordon2020compressingbertstudyingeffects} utilized \textit{k-means} clustering to identify shared weight representations for each layer within a trained NN. Therefore, weights falling into the same cluster were assigned a shared value. Changing bit precision affects the trade-off between model size and accuracy. Lower bit precision (e.g. FP16) allows the model size to be reduced, but may cause a loss of accuracy. Another quantization schemes involves INT8, in which both weights and activations are represented as 8-bit integers \cite{jacob2017quantizationtrainingneuralnetworks}. This technique is widely adopted for integer-arithmetic-only inference, enabling acceleration on CPUs and GPUs. 
Models with even lower precision include \textit{Ternary Weight Networks} \cite{li2022ternaryweightnetworks}, which quantize weights to $\{-1, 0, +1\}$. To mitigate the resulting loss of accuracy, \textit{quantization-aware training} anticipates inference-time quantization during training: it simulates low-precision behavior in the forward pass, while retaining full precision in the backward pass to preserve reliable gradient updates. This strategy maintains model accuracy while preparing DGMs for efficient deployment on resource-constrained IoT devices.
Both post-training quantization and quantization-aware training rely on access to the training dataset to achieve optimal performance, which may not always be feasible in privacy-sensitive scenarios, typical of some IoT domains. To overcome this limitation, \textit{Data-free quantization} \cite{nagel2019datafreequantizationweightequalization} has been introduced. This technique allow for the reduction of bit precision without requiring access to training data, offering a solution for privacy-constrained applications. Even with the most advanced quantization techniques, the maximum model compression ratio is inherently constrained by the smallest achievable bit width. Therefore, to further reduce model size and enable deployment over IoT devices, it is essential to integrate quantization methods with other compression strategies. However, quantization-supported techniques have been used in IoT applications for compressing DNNs \cite{9691274, 9442600} and DGMs \cite{10596048, lai2024ondemandquantizationgreenfederated}.

\paragraph{\textbf{Knowledge distillation (KD)}}
KD \cite{hinton2015distilling, buciluǎ2006model}
is a methodology that facilitates the transfer of knowledge from a complex model, referred to as the \textit{teacher model}, to a more compact model, the \textit{student model}. KD techniques generally necessitate training or fine-tuning to achieve effective knowledge transfer, thereby increasing the computational cost associated with their implementation. 
KD can be categorized into \textit{white-box} and \textit{black-box} approaches, depending on the level of access to the teacher model. 
\textit{White-box KD} provides access to the teacher’s architecture and internal representations, enabling the student to learn both outputs and intermediate features through feature-based losses. This structured guidance improves convergence and preserves key generative properties such as style and content consistency. \textit{Black-box KD} treats the teacher as an opaque model, training the student only on its final outputs. It is suitable when the teacher is proprietary or computationally expensive, but typically yields less accurate feature replication and may require more data.
KD reduces model complexity while maintaining performance, making it well suited for resource-constrained and edge environments. However, limited research has explored KD for DGMs in IoT contexts, and existing frameworks often lack the generalization needed for dynamic, heterogeneous IoT environments.

\paragraph{\textbf{Automated design: Compression and Neural Architecture Search}}
Model compression techniques rely heavily on hand-crafted heuristics and domain expertise to balance trade-offs among size, latency, energy, and accuracy. Because parameter sensitivity varies across layers, compression must be customized, making manual approaches time-consuming and often suboptimal. To overcome this, automated methods such as \textit{Automated Pruning} \cite{he2018amc} and \textit{Automated Quantization} \cite{wang2019haqhardwareawareautomatedquantization} use Reinforcement Learning (RL) to efficiently explore the design space and determine optimal pruning levels or bit-widths per layer.
In the context of IoT, where device capabilities, network conditions, and application requirements may change frequently over time, relying on static, manually crafted designs becomes particularly limiting. A similar need for automation therefore arises in the design of NNs themselves. Designing NNs is inherently complex and has traditionally depended on expert intuition and trial-and-error. To address this challenge, \textit{Neural Architecture Search (NAS)} \cite{zoph2017neuralarchitecturesearchreinforcement} extends the idea of automated design beyond compression by systematically identifying high-performing architectures without human intervention. NAS typically consists of three components: \textit{(i)} the search space, defining candidate architectures; \textit{(ii)} the search strategy, exploring them; and \textit{(iii)} the performance estimation, evaluating each configuration. When combined with model compression, NAS can be leveraged to automatically generate and adapt IoT-aware DGMs that are specialized for heterogeneous hardware profiles and evolving resource constraints, significantly accelerating deployment and reconfiguration without requiring continuous human intervention.  

\subsection{On device training, fine-tuning and inference}
Various techniques have been proposed to optimize the training phase directly on the devices, towards the goal of efficient on-device learning \cite{menghani2023efficient, li2023breaking}. One such method is \textit{gradient checkpoint}, designed to reduce the memory required for training activations by removing intermediate activations during the forward pass. These discarded activations are recomputed during the backward pass to calculate the gradients. Another approach, \textit{activation pruning}, reduces the size of the activation by removing non-critical neurons, similar to weight pruning, thus reducing the memory footprint and computational costs. \textit{Low-bit training}, on the other hand, involves training with quantizing weights, activations, and gradients, significantly reducing training cost. The techniques mentioned above primarily focus on scenarios in which NNs are trained from scratch, assuming a sufficient number of data samples. However, in on-device learning contexts, where data availability is limited, training DGMs from scratch becomes challenging due to the small dataset size. An alternative approach is to transfer pre-trained DGMs to the target device. This strategy leverages the advantages of existing, well-designed pre-trained DGMs that have been developed with extensive human expertise and significant computational resources, and adapt these to resource-constrained environments. Nevertheless, in dynamic IoT environments, newly collected data often deviates from previously learned distributions. Consequently, fine-tuning is frequently required to adapt DGMs effectively to new unseen data. Fine-tuning techniques can be broadly classified in: Parameter-Efficient Fine-Tuning \textit{(PEFT)} \cite{ding2023parameter}, \textit{Memory-Efficient Fine-Tuning (MEFT)} \cite{liao2023makepretrainedmodelreversible}, and \textit{Data-Efficient Fine-Tuning (DEFT)} 
\cite{zha2023datacentricaiperspectiveschallenges}. A summary of the proposed techniques is provided by Tab. \ref{tab:solutions}.
\begin{itemize}
    \item \textit{Parameter-Efficient Fine-Tuning}: PEFT reduces the computational cost of fine-tuning by selecting only a subset of essential parameters in DGMs for tuning. PEFT methods can be further divided in: \textit{(i)} \textit{Addition-Based Approach}, which introduces small neural network modules into the DGMs; \textit{(ii)} \textit{Specification-Based Approach}, that designates a sub-set of parameters for fine-tuning while freezing the rest; \textit{(iii)} \textit{Reparameterization-Based Approach}, that transforms the weight matrices into more efficient forms through LoRA. While PEFT significantly lowers computational costs, it still imposes a considerable memory footprint, restricting its use in IoT environments.
    \item \textit{Memory-Efficient Fine-Tuning}: MEFT minimizes the memory footprint during fine-tuning by employing various strategies as: \textit{(i)} avoiding storage of large input vectors, \textit{(ii)} utilizing low-energy optimizers, \textit{(iii)} combining gradient computation and update operations. MEFT offers a lower memory requirement compared to PEFT, but may take longer to complete and more energy consuming, which largely limit its use on IoT devices.
    \item \textit{Data-Efficient Fine-Tuning}: DEFT focuses on achieving efficient fine-tuning by utilizing only a small fraction of the data. These techniques are often integrated with PEFT or MEFT approaches to enhance fine-tuning efficiency, particularly in scenarios where data is scarce, making DEFT suitable for IoT applications.
\end{itemize}
To reduce the computational and memory footprint during inference, various techniques based on preprocessing input data have been proposed. Specifically, Chevalier et al. \cite{chevalier2023adaptinglanguagemodelscompress} utilize pre-trained LLMs to compress prompts with long contexts into shorter summary vectors, effectively reducing the overall computation and memory requirements. From a hardware optimization perspective, cross-processor inference has been introduced to enhance on-device inference efficiency. This approach involves distributing the modules of DGMs across multiple onboard processors, allowing parallel execution to improve efficiency. However, while general task-parallelism strategies have the potential to enhance performance, they are predominantly designed for server-side inference environments with homogeneous computational units. Finally, in IoT ecosystems, the available resources at runtime can be highly dynamic due to various factors such as device heterogeneity, energy constraints, and network variability. Consequently, the configuration of DGMs must be dynamically adjusted to adapt to these fluctuating resource conditions in real-time, ensuring effective and efficient on-device inference. Only a limited number of studies have explored the adaptation of DGMs, such as the work by \textit{Sheng et al.} \cite{sheng2023flexgenhighthroughputgenerativeinference}, which proposes techniques to tailor DGMs to various hardware resources. However, these approaches are primarily designed for resource-rich, server-scale systems. 

\paragraph{\textbf{Multi-device systems: Federated Learning (FL) approach}}
FL, despite its extensive study and application in recent years, has primarily been implemented in scenarios involving models of significantly smaller scales compared to contemporary DGMs, which are often characterized by billions of parameters. This highlights a gap in the current state of FL research and applications. Importantly, FL has the potential to address two critical challenges effectively: 
\begin{itemize}
    \item \textit{Privacy Preservation}: FL facilitates a decentralized training approach, ensuring that sensitive data remains local on IoT devices. This method minimizes the risk of data breaches and complies with strict privacy regulations by avoiding the need to transfer raw data to centralized servers.
    \item \textit{Scalable and Distributed Model Training}: By enabling multiple devices to collaboratively train smaller fractions of a large-scale model, FL offers a solution to the computational and communication challenges inherent in training massive DGMs. This approach not only optimizes resource utilization but also ensures that diverse, distributed data sources can contribute to model development without necessitating data centralization.
    To this end, \textit{Wen et al.} \cite{wen2022federateddropoutsimple} propose a simple approach to enable FL on resource-constrained devices. \textit{Alam et al.} \cite{alam2023fedrolexmodelheterogeneousfederatedlearning} propose FedRolex, a partial training-based approach that enables model-heterogeneous FL, and can train a global server model larger than the largest client model. Lastly, \textit{Dun et al.} \cite{dun2022efficientlightweightfederatedlearning} propose a novel asynchronous FL framework that utilizes dropout regularization to handle IoT device heterogeneity in distributed settings. 
\end{itemize}
So far, FL has been primarily presented as a solution to unlock the potential of GenAI within the IoT ecosystem. This synergy emphasizes FL's role in enabling privacy-preserving and distributed training for large-scale GenAI models, which often rely on data from multiple edge devices. However, the relationship between FL and GenAI is not unidirectional. GenAI techniques can also play a transformative role in enhancing the performance of current FL methodologies. As an example, \textit{Zhang et al.} \cite{zhang2024gptflgenerativepretrainedmodelassisted} propose a GPT-FL, a generative pre-trained model-assisted FL framework. 
To translate these theoretical advantages into real-world deployments, the gap between model size and edge constraints is currently being bridged by a specialized software stack. Optimized inference engines such as TensorFlow Lite Micro, ONNX Runtime, and NVIDIA TensorRT, provide essential quantization and pruning capabilities to fit DGMs onto constrained devices~\cite{ratul2025accelerating,zhou2023tensorrt}. Also, libraries specifically targeting LLMs at the edge, such as llama.cpp and MLC LLM, have emerged to enable 4-bit quantization on consumer-grade hardware~\cite{chen2025inference}.

\subsection{\textbf{Offloading}}
Given the constrained memory and computational capabilities of IoT devices, many of them may be unable to execute the most efficient DGMs, even when leveraging all the optimization techniques introduced thus far. In such scenarios, it becomes essential to offload the execution of the entire model or specific portions of it to external resources. However, the success of this offloading approach hinges on addressing two critical challenges: \textit{(i) efficient workload partitioning} and \textit{(ii) efficient communication}. These challenges become significantly more difficult to address in contexts involving DGMs due to their large-scale dimensions, which amplify the complexity of workload partitioning and place an even greater burden on communication efficiency. A summary is then provided by Tab. \ref{tab:solutions}.

\emph{Efficient workload partitioning.}
Workload Partitioning refers to the division of a DGM's computational tasks between resource-constrained IoT devices and nearby resource-rich devices as edge servers or cloud infrastructure. This enables the distributed execution of the model, leveraging the strengths of each system. However, this process is inherently complex due to the varying computational, memory, and energy capabilities of IoT devices, edge servers, and cloud resources. Existing approaches, to workload partitioning, can be categorized into two main types: \textit{(i) Heuristic-Based methods} that rely on predefined rules or experience-driven strategies to partition workloads and \textit{ (ii) Learning-Based methods}, which use historical workload data to train models that identify patterns and relationships between tasks and resources, enabling them to determine optimal partitions for new, unseen scenarios.\\
\emph{Efficient communication.}
Communication between IoT devices and cloud infrastructure is typically conducted over wireless channels, where bandwidth is often limited. Ensuring the timely exchange of offloaded workloads while minimizing bandwidth usage and power consumption is therefore critical. To address these challenges, several techniques have been proposed, including: \textit{(i) Message Compression} (i.e., reducing the size of transmitted data to conserve bandwidth), \textit{(ii) Data Sampling} (i.e., selecting representative data points to minimize the volume of communication), \textit{(iii) Efficient Communication Protocols} (i.e., optimizing data transmission methods for resource-constrained environments), and  \textit{(iv) Edge Caching} (i.e., storing frequently used data at edge servers to reduce repeated transmissions).

\subsection{\textbf{Solutions for Robust Deployments}}

In parallel to the design solutions discussed so far, the deployment of IoT systems is undergoing a paradigm shift towards High-performance embedded computing (HPEC), aimed at bridging the gap between cloud and edge by utilizing powerful accelerators, such as the NVIDIA Jetson Orin family, or dedicated Neural Processing Units (NPUs) integrated into modern SoCs. Unlike standard IoT nodes, these platforms are engineered to support energy-intensive matrix operations, enabling the local execution of DGMs while maintaining millisecond-level latency \cite{Navardi_Aalishah_Fu_Lin_Li_Chen_Mohsenin_2025, electronics14244877}. However, the physical accessibility of edge nodes introduces severe security risks. Unlike cloud data centers, IoT devices are susceptible to direct attacks and are generally less protected than other kind of infrastructure. Consequently, the deployment framework must prioritize hardware-backed security, to prevent attacks where proprietary DGMs weights are stolen, or poisoning attacks where the model’s behavior is maliciously altered.

Relying on secured HPEC for addressing the computational hurdles, however, solves only one side of the deployment equation. Even with the adoption of powerful edge accelerators, the efficacy of a GenAI model is fundamentally bounded by the quality and quantity of the data used for its training or fine-tuning. Unlike general-purpose domains, that benefit from massive public repositories, the IoT landscape suffers from severe data scarcity and fragmentation. Real-world sensor data is often proprietary, sensitive, or in the case of fault-detection statistically rare. Thus, to construct balanced IoT datasets for supervised learning is not straightforward. To establish baselines, researchers frequently rely on a handful of standard open-source repositories, such as the Bot-IoT or NSL-KDD datasets for network intrusion, C-MAPSS for industrial predictive maintenance, and PAMAP2 or MHEALTH for wearable sensor data. However, these static datasets could fail to capture the dynamic, heterogeneous nature of modern, real-world IoT deployments.

To overcome these limitations, GenAI provides a powerful alternative by generating synthetic data that realistically simulates complex, evolving environments. Beyond simply augmenting sparse datasets, this generative capability addresses privacy and regulatory challenges associated with deploying AI in constrained environments. By leveraging synthetic datasets, practitioners can train downstream models on statistically representative data without exposing raw, personally identifiable information, thereby facilitating compliance with regulations in smart home or healthcare applications \cite{s24227389, electronics13173509}. This is increasingly achieved through specialized data generation tools and libraries, as: Synthetic Data Vault (SDV), DataSynthesizer or CTGAN, which are tailored to handle the complex, multivariate time-series data typical of IoT sensors. While synthetic data generation addresses privacy by removing personally identifiable information, it raises distinct ethical concerns regarding "bias amplification" \cite{wyllie2024fairness}. If a generative model is trained on a limited, non-representative sample of IoT data, it risks hallucinating synthetic datasets that reinforce existing biases. In healthcare or smart city applications, this can lead to automated decisions that are statistically skewed against specific demographics, directly eroding user trust. Therefore, data frameworks must now include bias-detection loops alongside standard validation~\cite{koul2025synthetic}. Furthermore, establishing standardized benchmarks remains a critical hurdle for robust deployment. While the IoT community utilizes established hardware-centric benchmarks like MLPerf Tiny or AIoTBench to evaluate the inference latency and energy efficiency of traditional discriminative models on edge accelerators, evaluating GenAI introduces new complexities. The industry currently lacks a unified benchmark suite that simultaneously assesses the edge‑hardware efficiency of a DGM and the statistical fidelity of its outputs. As a result, researchers often adapt traditional generative evaluation metrics to time‑series sensor data, underscoring the need for dedicated GenAI‑IoT evaluation frameworks.

\begin{table*}[hbt!]
\footnotesize	
\centering
\caption{Main Solutions for practical IoT challenges  }
\label{tab:solutions}
\renewcommand{\arraystretch}{1.3}
\resizebox{\textwidth}{!}{
\begin{tabular}{p{0.10\linewidth} p{0.25\linewidth} p{0.35\linewidth} p{0.35\linewidth}}
\toprule
\textbf{Technique} & \textbf{Objective} & \textbf{IoT Advantage} & \textbf{IoT Disadvantage} \\ \bottomrule

\multicolumn{4}{l}{\textit{\textbf{Model Compression \& Architecture Optimization}}} \\ \toprule

\textbf{Pruning} & 
Eliminates redundant parameters to induce model sparsity & 
\textbf{Reduced Footprint:} Lowers storage (Flash) and RAM usage \newline 
\textbf{Energy Savings:} Reduces memory access costs & 
\textbf{Hardware Dependency:} Unstructured pruning requires specialized accelerators for speedup \newline 
\textbf{Retraining Cost:} Requires fine-tuning\\ \midrule

\textbf{Quantization} & 
Maps high-precision weights and activations to lower-bit representations (e.g., INT8) & 
\textbf{Throughput:} Enables fast integer-only arithmetic on edge NPUs \newline 
\textbf{Efficiency:} Reduces memory bandwidth pressure & 
\textbf{Quality Degradation:} Sensitive to activation outliers (functional collapse) \newline 
\textbf{Hardware Fit:} Aggressive quantization ($<4$ bits) needs specific HW support \\ \midrule

\textbf{Knowledge Distillation} & 
Transfers knowledge from a large "teacher" to a compact "student" & 
\textbf{HW Compatibility:} Produces dense models optimized for standard CPUs
\newline 
\textbf{Inference Speed:} Decouples training size from runtime speed & 
\textbf{Capacity Gap:} Small models may suffer from mode collapse (less diversity) \newline 
\textbf{Training Overhead:} Requires access to the heavy teacher model \\ \bottomrule

\multicolumn{4}{l}{\textit{\textbf{Efficient Fine-Tuning (PEFT)}}} \\ \toprule

\textbf{LoRA} & 
Injects trainable low-rank matrices into frozen layers to approximate updates & 
\textbf{Adaptation:} Enables low-cost on-device personalization \newline 
\textbf{Modularity:} Minimal storage to switch between tasks & 
\textbf{Memory Constraint:} Unreduced runtime RAM usage of the base model \newline 
\textbf{Latency:} Slight overhead if adapters are not merged \\ \midrule

\textbf{PEFT \newline(General)} & 
Updates only a small subset of parameters (e.g., Adapters, Prefix Tuning) & 
\textbf{Storage Efficiency:} Drastically reduces storage needs for checkpoints \newline 
\textbf{Stability:} Prevents catastrophic forgetting & 
\textbf{Inference Latency:} Adapter layers introduce extra computation steps \newline 
\textbf{Base Model Dependency:} The heavy backbone still in memory \\ \bottomrule

\multicolumn{4}{l}{\textit{\textbf{Training \& Data Strategies}}} \\ \toprule

\textbf{MEFT} & 
Reduces peak memory usage during training (e.g., Gradient Checkpointing) & 
\textbf{Feasibility:} Allows training LLM on RAM-constrained devices \newline 
\textbf{Stability:} Prevents Out-Of-Memory crashes & 
\textbf{Computational-Energy Cost:} Recomputing activations increase CPU use \newline 
\textbf{Throughput:} Slows down the training process \\ \midrule

\textbf{DEFT} & 
Fine-tunes models using limited, high-value data samples & 
\textbf{Data Efficiency:} Reduces data collection and transmission costs \newline 
\textbf{Speed:} Faster training cycles with small datasets & 
\textbf{Overfitting Risk:} High susceptibility to local noise/bias \newline 
\textbf{Generalization:} Risk of fail on unseen scenarios \\ \bottomrule

\multicolumn{4}{l}{\textit{\textbf{Distributed \& Collaborative Intelligence}}} \\ \toprule

\textbf{Federated Learning} & 
Distributes training across devices; aggregates updates centrally & 
\textbf{Privacy:} Raw user data never leaves the edge device \newline 
\textbf{Scalability:} Leverages the collective compute of the IoT fleet & 
\textbf{Communication Cost:} Frequent updates via radio drain battery \newline 
\textbf{Heterogeneity:} Non-IID data causes training instability \\ \midrule

\textbf{Offloading} & 
Divides model execution layers between device and edge/cloud & 
\textbf{Memory Extension:} Enables inference of massive models exceeding local RAM \newline 
\textbf{Load Balancing:} Offloads heavy computation & 
\textbf{Network Dependency:} Latency fluctuates with bandwidth stability \newline 
\textbf{Privacy:} Intermediate data transmission expands attack surface \\ \midrule

\textbf{Efficient \newline communication} & 
Compresses gradients/activations before transmission & 
\textbf{Bandwidth Efficiency:} Reduces radio active time \newline 
\textbf{Throughput:} Accelerates distributed training & 
\textbf{Compute Overhead:} Compression algorithms consume CPU energy \newline 
\textbf{Accuracy Risk:} Lossy compression is error-prone\\ \bottomrule

\end{tabular}
}
\end{table*}

%% file: survey_sections/conclusion.tex
\section{Conclusions}
\label{sec:conclusion}
The question of how AI should 
support IoT  
opens a significant debate in current computer science and engineering. Over the years, Discriminative AI models have been instrumental in supporting the design and implementation of intelligent IoT applications: tools (such as ML and DL) effectively extract patterns and correlations from large datasets, being focused solely on partitioning and categorizing data points in order to produce a probabilistic classification or decision. In a broader perspective, the estimation of probability underpinning the GenAI can serve as a foundational element to comprehensively address inherent semantic-, networking- and things-related issues featuring the IoT, by
 simultaneously handling uncertainty and understanding and, hence, to generate outputs creatively. 

Such a research direction is still mostly unexplored, and motivated this systematic survey in providing both a quantitative and qualitative analysis of the body of knowledge related to the integration of GenAI within IoT Computing.  
As takeaways of this survey, we recognized that comprehensive and domain-independent secondary studies on the GenAI-IoT duo are very few, while,  
among the application-oriented primary studies, DGMs are widely employed in both their explicit and implicit density modeling forms. On the explicit density side, ARMs, particularly those leveraging Transformer-based architectures, are the most commonly adopted to process sequential sensor data, predict future states, and enhance automation (particularly, LLMs have primarily and largely been exploited for interpreting network traffic profile or  implementing natural language interfaces for Smart Devices, thus originating the conflation of conversational applications with GenAI). On the implicit density side, GANs dominate the IoT landscape due to their ability to generate high-fidelity data without requiring explicit likelihood estimation. Nevertheless, other generative approaches such as VAEs and Diffusion models also feature prominently in the literature. 

Despite the specific family of DGM, the majority of studies emphasize the challenges associated with deploying these models on resource-constrained devices, which remains a critical barrier to their full adoption in real-world IoT scenarios. In response to these limitations, various optimization techniques (ranging from model compression to edge-aware adaptations) have been outlined but not yet fully explored in IoT Computing. Conversely, a significant barrier remains the lack of standardized evaluation frameworks. While benchmarks like MLPerf Tiny exist for discriminative models, the industry lacks a unified standard to evaluate the hallucinations or energy efficiency of generative models in IoT loops. Currently, researchers often repurpose datasets such as NASA’s C-MAPSS for predictive maintenance to test GenAI robustness, but dedicated frameworks for validating synthetic data quality in safety-critical systems remain an open research challenge. Moreover, the "black box" nature of DGMs presents a critical challenge regarding Explainable AI (XAI). In safety-critical IoT loops, such as autonomous braking or industrial control, operators require more than just a prediction; they require interpretability but current edge frameworks often lack lightweight XAI tools.

To conclude, as articulated in this survey, we strongly believe that as IoT grows, GenAI will play a crucial role in scaling and managing its complexity, thus outlining 
an exciting and multidisciplinary area for future work. 